%% file: main.tex
\pdfoutput=1

\documentclass[11pt]{article}

\usepackage[preprint]{acl}

\usepackage{times}
\usepackage{latexsym}
\usepackage[T1]{fontenc}
\usepackage[utf8]{inputenc}
\usepackage{microtype}
\usepackage{inconsolata}
\usepackage{graphicx}

\usepackage{booktabs}
\usepackage{multirow}
\usepackage{makecell}
\usepackage{amsmath}
\usepackage{amssymb}
\usepackage{amsthm}
\usepackage[table]{xcolor}
\usepackage{pifont}
\usepackage{xspace}
\xspaceremoveexception{/}
\usepackage{placeins}
\usepackage{algorithm}
\usepackage{algpseudocode}
\usepackage[capitalise,noabbrev,nameinlink]{cleveref}

\renewcommand{\topfraction}{0.9}
\renewcommand{\dbltopfraction}{0.9}
\renewcommand{\textfraction}{0.07}
\renewcommand{\floatpagefraction}{0.85}
\renewcommand{\dblfloatpagefraction}{0.85}
\newtheorem{theorem}{Theorem}
\newtheorem{lemma}{Lemma}

\newcommand{\method}{LoopSkim\xspace}
\newcommand{\methodH}{LoopSkim~+~head\xspace}
\newcommand{\X}{\ensuremath{\times}\xspace}
\newcommand{\Tone}{\ensuremath{T{=}1}\xspace}
\newcommand{\cmark}{\ding{51}}
\newcommand{\xmark}{\ding{55}}
\definecolor{up}{HTML}{0072B2}
\definecolor{dn}{HTML}{A2457A}
\definecolor{nt}{HTML}{7F7F7F}
\definecolor{ours}{HTML}{FDEDE4}

\input{numbers}
\title{\method: Lossless Multi-Depth Speculative Decoding for Looped Language Models}

\author{Muhan Xiao$^{1}$ \quad Shijian Li$^{2,}$\thanks{Corresponding author.} \\
  $^{1}$College of Artificial Intelligence \\
  $^{2}$College of Computer Science and Technology \\
  Zhejiang University \\
  \texttt{\{humor,shijianli\}@zju.edu.cn}}

\begin{document}
\setcounter{footnote}{1}
\maketitle

\begin{abstract}
\input{sections/00_abstract}
\end{abstract}

\input{sections/fig_method}

\input{sections/01_introduction}
\input{sections/02_background}
\input{sections/03_method}
\input{sections/04_experiments}
\input{sections/05_conclusion}
\input{sections/06_limitations}

\nocite{zhu2025ouro}
\bibliography{references}

\clearpage

\appendix
\crefalias{section}{appendix}
\crefalias{subsection}{subappendix}
\let\appendixsection\section
\renewcommand{\section}{\FloatBarrier\appendixsection}

\input{sections/app_A_setting}
\input{sections/app_B_proof}
\input{sections/app_C_results}

\end{document}

%% file: numbers.tex
\newcommand{\nVsLoopSpec}{1.99}  %
\newcommand{\nVsCopyLoop}{1.51}  %
\newcommand{\nFreeVsLoopSpec}{1.73}  %
\newcommand{\nFreeVsCopyLoop}{1.30}  %
\newcommand{\nFPPrompts}{3,300}          %

\newcommand{\nFreeARGreedy}{7.66}  %
\newcommand{\nFreeARTone}{4.87}  %

\newcommand{\nHeldVsLoopSpec}{1.90}  %
\newcommand{\nEngineMin}{1.85}  %
\newcommand{\nEngineMax}{2.35}  %

\newcommand{\nLoopTimeMax}{1.27}  %

\newcommand{\nThinkStart}{1.35}  %
\newcommand{\nThinkEnd}{1.93}  %

\newcommand{\nGreedySignP}{0.96}          %
\newcommand{\nToneSignP}{0.24}           %
\newcommand{\nToneDiff}{\ensuremath{-0.22}}   %
\newcommand{\nToneDiffLo}{\ensuremath{-0.60}} %
\newcommand{\nToneDiffHi}{\ensuremath{+0.14}} %
\newcommand{\nGreedyDiff}{\ensuremath{-0.01}}   %
\newcommand{\nGreedyDiffLo}{\ensuremath{-0.29}} %
\newcommand{\nGreedyDiffHi}{\ensuremath{+0.26}} %
\newcommand{\nBFRefP}{0.50}                %

\newcommand{\nFreeBootLo}{1.71}  %
\newcommand{\nFreeBootHi}{1.74}  %
\newcommand{\nBootLo}{1.97}  %
\newcommand{\nBootHi}{2.00}  %
\newcommand{\nFreeNoRep}{1.62}   %
\newcommand{\nNoRep}{1.88}   %
\newcommand{\nFreeHeldVsLoopSpec}{1.69}  %
\newcommand{\nGPUFreeMin}{1.36}  %
\newcommand{\nGPUFreeMax}{2.13}  %
\newcommand{\nGPUMin}{1.57}  %
\newcommand{\nGPUMax}{2.78}  %
\newcommand{\nAdmRej}{23}  %
\newcommand{\nInterShare}{94}  %
\newcommand{\nFullRate}{0.43}  %
\newcommand{\nInterShareTone}{87}  %
\newcommand{\nIndepToneOuro}{0.764}  %
\newcommand{\nIndepToneRaven}{0.331}  %
\newcommand{\nLadXNOuro}{0.888}  %
\newcommand{\nLadXNRaven}{0.754}  %
\newcommand{\nLadXtwoN}{0.552}  %
\newcommand{\nLadXNToneOuro}{0.862}  %
\newcommand{\nLadXNToneRaven}{0.792}  %
\newcommand{\nMeanDepRvl}{6.7}  %
\newcommand{\nWasteRaven}{16}  %
\newcommand{\nWasteOuro}{48--50}  %
\newcommand{\nKeptFirst}{94}  %
\newcommand{\nBFOurs}{75.5}  %
\newcommand{\nBFLS}{70.1}
\newcommand{\nBFCL}{70.6}
\newcommand{\nBFAR}{68.9}  %
\newcommand{\nGPUFreeCLMin}{0.96}  %
\newcommand{\nGPUFreeCLMax}{1.22}  %
\newcommand{\nHugFreeAR}{7.42}  %
\newcommand{\nHugHeadAR}{8.24}  %
\newcommand{\nHugFreeCL}{1.29}  %
\newcommand{\nHugFreeLS}{1.98}  %
\newcommand{\nHugFreeLSNoRep}{1.59}  %
\newcommand{\nEngineFreeMin}{1.64}  %
\newcommand{\nEngineFreeMax}{2.08}  %
\newcommand{\nLadXNMin}{0.73}  %
\newcommand{\nLadXNMax}{0.92}  %

%% file: sections/00_abstract.tex
Looped language models such as Ouro and Raven apply one weight-shared block several times per token, so autoregressive decoding (AR) runs every loop of every token in sequence.
Pipelined self-speculative decoders overlap these loops by drafting tokens from shallow readouts and verifying them at full depth.
Yet every loop of a looped LM ends in a readout of the next token, and draft tokens pass through every depth anyway, so each depth can check them without an extra pass.
We introduce \method, in which the depth of the pipeline is the verifier. Draft tokens copied from the context, as in prompt lookup decoding, enter after the first loops, any later readout that contradicts one replaces it at that depth, and only full-depth readouts commit tokens.
\method fixes the random draws of every position in advance, so each check compares two tokens and decoding stays lossless in exact arithmetic under sampling for any number of checks and draft chains.
Without training, its speedup over AR averages \nFreeVsLoopSpec\X that of LoopSpec across greedy decoding and sampling, and its checks between the first loops and full depth speed it up on every model.
An optional draft head adds a second chain and raises this average to \nVsLoopSpec\X.

%% file: sections/fig_method.tex
\begin{figure*}[t]
  \centering
  \includegraphics[width=\textwidth]{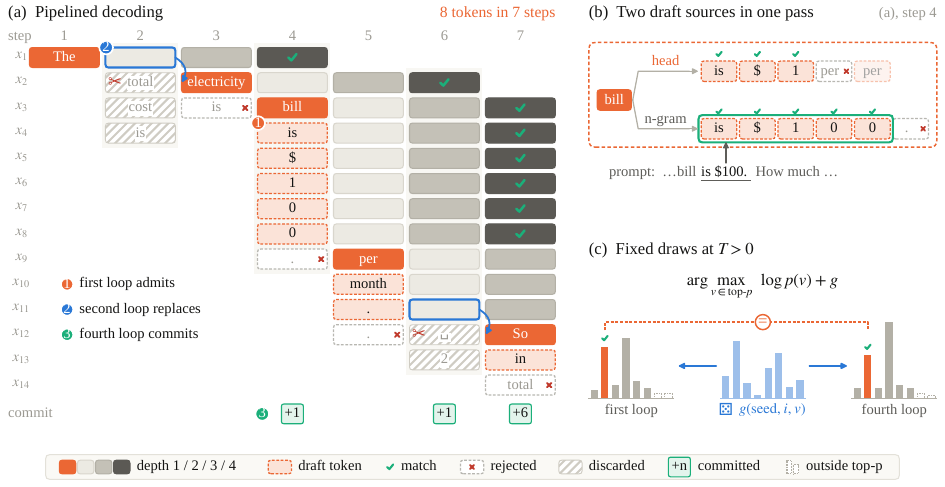}
  \caption{(a)~A greedy decode of Ouro-1.4B ($R=4$) with both draft chains: 8 tokens take 7 one-loop steps, against $8 \times 4 = 32$ loops for AR.
    (b)~In step~4, the n-gram chain has more admitted tokens than the head's and is kept.
    (c)~At $T>0$, all readouts at position $i$ add the same Gumbel noise $g$, so each check compares two tokens.}
  \label{fig:method}
\end{figure*}

%% file: sections/01_introduction.tex
\section{Introduction}\label{sec:intro}

Looped language models (looped LMs) apply one weight-shared block of layers several times per token \citep{dehghani2019universal}, so they compute more without storing more parameters.
Ouro models of at most 2.6B parameters match LLMs of up to 12B \citep{zhu2025ouro}, and Raven models reason better as they run more loops at test time \citep{geiping2025huginn,mcleish2025retrofitted}.
This depth costs latency. Autoregressive decoding (AR) runs all $R$ loops of a token before the next token starts, although at small batch sizes a forward pass over several positions costs far less than one pass per position.

Pipelined self-speculative decoding \citep{hooper2023speed,cho2026loopspec,ha2026wavefront} fills this spare capacity. A shallow readout applies the output layers after the first loops. A token drafted from it starts its loops while earlier tokens finish theirs, so one pipeline step advances positions at different depths.
These decoders face a trade-off. LoopSpec and WaveFront draft only one token ahead, so a step commits at most one token. Drafting several tokens per step, as concurrent decoders do \citep{liu2026loopahead,du2026copyloop}, lifts this cap but makes a wrong draft token costly.
If only full depth finds it, the pipeline restarts from the corrected token and discards the loops of every token behind it.

A looped LM already holds the verifier these draft tokens need: its own depth. Ouro is trained to read out the next token after every loop \citep{zhu2025ouro} and Raven after a random number of loops \citep{geiping2025huginn}, and draft tokens pass every depth on their way to full depth anyway, so each depth can check them without an extra pass.
In \method (\cref{fig:method}), draft tokens copied from the context, as in prompt lookup decoding \citep{saxena2023prompt}, run their first loops with the newest token, and the readouts at this depth, the admission check, admit those they reproduce.
Every later check depth, each loop on Ouro and each doubling of the loop count on Raven, compares its readouts with the admitted tokens, replaces the first one it contradicts and restarts the pipeline from there. Only full-depth readouts commit tokens. We call this multi-depth speculation.
On Raven-Llama under greedy decoding, a replacement after the admission check comes after a mean of \nMeanDepRvl{} of its 32 loops, so the positions behind a wrong token stop long before full depth.

Under sampling, speculative sampling at each of these depths \citep{leviathan2023fast,chen2023accelerating} would carry a residual distribution from each replaced token into every later check. \method instead fixes the random draws of every position in advance (\cref{sec:method-noise}), so every readout of a position draws with the same Gumbel noise and the full-depth draw depends only on the seed and the prefix.
Admitting, replacing and redrafting tokens then change what is computed ahead of time, not what is committed, so in exact arithmetic \method is lossless for any number of checks and draft chains (\cref{thm:lossless}).
\method thus drafts several tokens without training, checks them at every check depth, and stays lossless under sampling. An optional one-layer draft head adds a second chain where training data exist. Our contributions are:
\begin{itemize}
\item We use a looped LM's depth as the verifier of its own pipeline (multi-depth speculation). Each check depth replaces the first draft token its readouts contradict, so the tokens behind it stop before full depth. \method needs no training and no extra pass (\cref{sec:method-cycle}).
\item We fix the random draws of every position in advance, as predictive sampling does, and Raven's random initial states. This keeps \method lossless under sampling, and each check compares two tokens, which usually agree when the two readouts are close (\cref{sec:method-noise,sec:method-sources}).
\item We evaluate \method on five Ouro and Raven models (\cref{sec:experiments}). Without training, its speedup averages \nFreeVsLoopSpec\X that of LoopSpec and is \nEngineFreeMin--\nEngineFreeMax\X that of LoopSpec's schedule in our engine. Its checks between admission and full depth speed it up on every model.
\end{itemize}

%% file: sections/02_background.tex
\section{Background and Related Work}\label{sec:bg}

\subsection{Looped LMs and the Speedup Equation}\label{sec:background}\label{sec:bg-speedup}

One application of the weight-shared block of layers is a loop, which Raven calls a recurrent step, and a position's recurrent depth $\ell$, or depth, counts its loops.
Applying the output layers to the state $h^{(\ell)}_{i-1}$ gives the readout $p_\ell(\cdot \mid x_{<i})$, with $p_R$ the model's distribution. The readout token $\hat{x}^{(\ell)}_i$ is its arg max under greedy decoding and a draw with the fixed noise of \cref{sec:method-noise} under sampling.
Ouro \citep{zhu2025ouro} loops its whole layer stack. Raven \citep{geiping2025huginn,mcleish2025retrofitted} loops a recurrent block between a prelude and a coda, starting each position's state from random noise.
As in LoopSpec, every token runs all $R$ loops, and each loop keeps its own KV cache.

In a pipeline, each active token position, or position, is at one depth. It takes the token at that position as input, and its readout predicts the next token.
A pipeline step, or step, is one forward pass that advances every active position by $d$ loops, the stride. A new token and its draft tokens run their first step together, so their admission check reads out at depth $\ell_1 = d$.
As earlier positions are never shallower than later ones, every KV entry a position reads already exists.

With $\tau$ tokens committed per step on average (the counterpart of the mean accepted length in speculative decoding), the sequential loops per output token relative to AR, $D$, and the speedup are
\begin{equation}
  D = \frac{d}{R\,\tau}, \qquad
  \text{speedup} = \frac{1}{D} \cdot \frac{t_{\mathrm{AR}}}{t_{\mathrm{loop}}},
  \label{eq:speedup}
\end{equation}
where $t_{\mathrm{AR}}$ is AR's decode time per loop and $t_{\mathrm{loop}}$ the pipeline's time per step, host work included, divided by $d$.
Admitting one new position per step caps $\tau$ at one, so $D \ge d/R$. Admitting $G$ draft tokens for the following positions lifts the cap to $1+G$.
The cost is wider passes. A group is the positions admitted in one step, at most $1+G$. With $C$ draft chains of $G$ tokens at the admission check and $R/d-1$ deeper active groups, a pass carries up to $(C + R/d - 1)(1+G)$ tokens where AR carries one, which raises $t_{\mathrm{loop}}$ (\cref{sec:ablations}).

\subsection{Related Work}\label{sec:related}

\paragraph{Efficient decoding of looped LMs.}
SPEED \citep{hooper2023speed} pipelines single-token drafts through cyclically shared decoder layers but needs custom training and is evaluated only greedily.
Huginn \citep{geiping2025huginn} drafts tokens with fewer loops of the same model and verifies them at full depth.
LoopSpec \citep{cho2026loopspec} drafts a token from a shallow readout, adds a second proposal from a deeper readout when a gate opens, and verifies them in turn at full depth by speculative sampling.
WaveFront \citep{ha2026wavefront} pipelines single-token drafts and DAS-Wave \citep{li2026das} verifies drafted blocks in a separate full-depth pass, each evaluated greedily.
Concurrently, LoopAhead \citep{liu2026loopahead} drafts a tree of positions with a trained drafter and pauses those that its intermediate checks contradict, and CopyLoop \citep{du2026copyloop} checks copied drafts at a shallow depth and verifies them at full depth, evaluated under greedy decoding (\cref{tab:design}).
Other work updates loops lazily or shares KV caches across them \citep{yang2026flashloop,prairie2026benice}, trains loops to run concurrently \citep{wu2025plt,hao2026dualslot}, as LoopFold does \citep{song2026loopfold}, or, without retraining, lets earlier tokens keep refining while each pass decodes a new one \citep{geiping2025parallel}.
Adaptive computation runs fewer loops for some tokens, by learned halting \citep{graves2016adaptive,banino2021pondernet}, early exit \citep{schuster2022calm,bae2023free,bae2025relaxed}, routed recursion depths \citep{bae2025mor} or the per-token exits of Ouro and Huginn \citep{zhu2025ouro,geiping2025huginn}, and so changes the output, unlike \method.

\paragraph{Speculative decoding.}
Speculative decoding \citep{leviathan2023fast,chen2023accelerating} takes drafts from the target's own skipped or early-exited layers \citep{zhang2024draft,elhoushi2024layerskip,liu2024kangaroo}, from heads on its features \citep{cai2024medusa,li2024eagle}, or, without training, from n-grams of the context and Jacobi iterations \citep{saxena2023prompt,oliaro2025suffix,song2021accelerating,santilli2023accelerating,fu2024lookahead}.
Multi-draft verification keeps sampling lossless by testing further drafts against residual distributions \citep{miao2024specinfer,sun2023spectr}.
Staged and hierarchical schemes let a cheaper model check drafts before the target does \citep{spector2023staged,chen2024cascade,sun2024triforce,kumar2025hispec,mcdanel2025pipespec}, and each such check runs its own pass, through a smaller model or the target's first layers. The checks of \method instead read out draft tokens at depths they pass anyway.

\paragraph{Coupled sampling.}
Predictive sampling \citep{wiggers2020predictive}, drafter-invariant speculative decoding \citep{daliri2024coupling} and list-level coupling for several drafts \citep{rowan2025coupling} share random numbers between distributions so that their draws often coincide, and parallel samplers of autoregressive models fix the noise of every position in advance \citep{song2021accelerating}.
\method draws every readout of a position with the same noise and fixes Raven's initial states in the same way (\cref{sec:method-noise}).

%% file: sections/03_method.tex
\section{Method}\label{sec:method}

\method lowers $D$ by raising $\tau$, the tokens committed per step. Draft tokens enter after the first loops, every later check depth can replace them, and only full depth commits.

\subsection{Multi-Depth Speculation}\label{sec:method-cycle}

In step~4 of \cref{fig:method}a, the newest token is ``bill''. The n-gram chain copies ``is \$ 1 0 0 .''\ (one token per digit) from the question.
The admission check reproduces the first five of these draft tokens and reads ``per'' where the chain drafted ``.'', so ``per'' becomes the newest token.

In general the newest token is $x_P$, and the active positions before $P$ form groups, each at one depth.
Checks run at the depths $\Lambda = \{\ell_1 < \ell_2 < \dots < R\}$, a subset of the multiples of $d$, from the admission check at $\ell_1 = d$.
In each step, a draft chain proposes draft tokens $u_1, \dots, u_G$ for positions $P+1, \dots, P+G$, and they run their first step together with $x_P$.

\paragraph{Admit.}
Position $P+j-1$ has read exactly the tokens before $P+j$, so its readout token $\hat{x}^{(\ell_1)}_{P+j}$ at the admission check is what depth $\ell_1$ would place there and checks the draft token $u_j$.
The admitted draft tokens are the longest run that these readouts reproduce,
\begin{equation}
  a = \max\bigl\{ 0 \le j \le G : u_r = \hat{x}^{(\ell_1)}_{P+r} \;\forall\, 1 {\le} r {\le} j \bigr\}.
  \label{eq:admit}
\end{equation}
Positions $P, \dots, P+a$ enter as one group, the readout token $\hat{x}^{(\ell_1)}_{P+a+1}$ becomes the newest token, and the remaining draft tokens are discarded.

\paragraph{Check and replace.}
When a group of $n$ positions $P_0, \dots, P_0+n-1$ reaches a check depth $\ell \in \Lambda$ with $\ell_1 < \ell < R$, the readout at each position is compared with the token that follows it:
\begin{equation}
  m = \min\bigl\{ 1 \le j \le n : x_{P_0+j} \ne \hat{x}^{(\ell)}_{P_0+j} \bigr\}.
  \label{eq:trunc}
\end{equation}
For $j = n$, the token that follows is the first of the next group, or the newest token.
If such an $m$ exists, $\hat{x}^{(\ell)}_{P_0+m}$ replaces $x_{P_0+m}$, every position $\ge P_0+m$ is dropped, later groups included with their readouts in this pass, and the pipeline restarts from the corrected token ($P \gets P_0+m$).
A wrong token is thus replaced at the first check depth whose readout contradicts it, instead of at $R$, and the positions behind it restart there.
In step~2 of \cref{fig:method}a, the second loop reads ``electricity'' after ``The'' where the first had read ``total'', so the tokens from ``total'' on are dropped and ``electricity'' enters the pipeline in step~3.
A step handles its readouts from the deepest group to the admission check, and the first replacement ends the step before any draft token is admitted.

\paragraph{Commit.}
A group's positions reach depth $R$ together, and \cref{eq:trunc} at $\ell = R$ commits, in position order, each token that their readouts reproduce, up to the first mismatch, plus the full-depth readout token that replaces it. The mismatch drops every uncommitted token, and the pipeline restarts from that token.

\subsection{Lossless Sampling}\label{sec:method-noise}

\paragraph{Greedy decoding.}
Every check in the pipeline compares a token at position $i$, a draft token or an earlier readout token, with the readout token $\hat{x}^{(\ell)}_i$ of some depth $\ell$.
Under greedy decoding $\hat{x}^{(\ell)}_i$ is the arg max of $p_\ell(\cdot \mid x_{<i})$, so a check compares two tokens, and $\hat{x}^{(R)}_i$ is AR's token.

\paragraph{Fixed draws.}
At $T>0$ readout tokens are random draws, and independent draws at two depths often disagree even for equal distributions.
\method fixes the draws in advance (\cref{fig:method}c).
For seed $s$, every position $i$ and token $v$ carry an i.i.d.\ Gumbel(0,\,1) noise $g_{s,i,v}$, and every readout, at any depth and for any draft chain, draws
\begin{equation}
  \hat{x}_i^{(\ell)} = \operatorname*{arg\,max}_{v \in \mathcal{V}_p^{(\ell)}}
  \bigl[\log p_\ell(v \mid x_{<i}) + g_{s,i,v}\bigr],
  \label{eq:shared-noise}
\end{equation}
where $p_\ell$ is the depth-$\ell$ readout at temperature $T$ and $\mathcal{V}_p^{(\ell)}$ its top-$p$ set. Greedy decoding is the case $g \equiv 0$.
By the Gumbel-max trick \citep{gumbel1954statistical,maddison2014astar}, $\hat{x}_i^{(R)}$ is a sample from the model's top-$p$ distribution, so AR with this noise is ordinary top-$p$ sampling.

The fixed draws serve two purposes.
At full depth, they make the token at position $i$ a function of the seed and the prefix alone, so every check compares a draft token with this function's value, and sampling runs the greedy algorithm unchanged. This makes \method lossless.
Below full depth, a readout that reuses the noise of its position draws the same token as full depth whenever the perturbed arg max coincides, which close distributions make likely \citep{wiggers2020predictive,daliri2024coupling}. This keeps intermediate checks accurate.
As only full-depth draws are committed, readouts below full depth and draft chains may use any temperature or truncation, and \method prunes those readouts by a margin and lets the head draw at a lower temperature (\cref{tab:app-settings}).

\paragraph{Random initial states.}
For Raven (\cref{sec:background}), \method derives the initial state $\xi_{s,i}$ from seed, position and hidden dimension, so AR, every draft and every restart start position $i$ from the same state, drawn from the released model's initial-state distribution (\cref{app:proof-raven}).

\begin{theorem}[Lossless decoding]\label{thm:lossless}
  In exact arithmetic, for every prompt and seed $s$, after every pipeline step the committed tokens of \method are a prefix of the tokens that AR generates with the same noise $g_{s,\cdot,\cdot}$ and, on Raven, the same initial states $\xi_{s,\cdot}$, whatever draft tokens the steps admit, replace or discard. At termination, the two outputs are equal.
  Hence \method returns the model's greedy output under greedy decoding and, at $T>0$, samples exactly from the model's top-$p$ distribution.
\end{theorem}

\paragraph{Proof sketch.}
The proof (\cref{app:proof}, \cref{lem:reads}) rests on three properties of a pipeline step.
(i)~Every state that a position reads, cached or computed in the same pass, equals what AR computes from the current tokens up to its position.
(ii)~When a token changes, every state at its position or later is invalidated, while earlier states stay.
(iii)~A token is committed only from the full-depth readout at a position whose input tokens are all committed.
If the committed tokens $x_{\le c}$ equal AR's, then by (iii) the next one is the full-depth readout token at position $c$, which by (i) and (ii) reads AR's states and by \cref{eq:shared-noise} uses AR's noise $g_{s,c+1,\cdot}$. It is AR's $x_{c+1}$, and induction completes the proof.

\subsection{Draft Chains}\label{sec:method-sources}\label{sec:method-guesses}

\paragraph{N-gram chain.}
By default, \method copies one chain of draft tokens from the context, as in prompt lookup decoding \citep{saxena2023prompt}. The n-gram chain finds the last earlier occurrence of the longest suffix ending at the newest token, trying four tokens down to two, and copies up to $G$ tokens that followed it in the prompt or generated text.
The lookup runs on the host and needs no forward pass or training.

\paragraph{Optional draft head.}
By \cref{thm:lossless}, no draft chain can change the output, so \method can run $C$ chains at once. Each chain runs its first step on $x_P$ in the same pass, \cref{eq:admit} gives the admitted tokens $a_k$ of chain $k$, and the chain with the largest $a_k$ is kept (\cref{fig:method}b).
Where training data are available, a one-layer draft head on the frozen model adds a second chain, like drafting heads on a model's features \citep{cai2024medusa,li2024eagle,ankner2024hydra} and LoopAhead's drafter on early-loop states \citep{liu2026loopahead}.
It reads the depth-$\ell_1$ states of a recent window and predicts the depth-$\ell_1$ states of the next $G$ positions in one pass.
Its draft tokens first meet the admission check, so it is trained toward $p_{\ell_1}$ on the model's own continuations rather than toward $p_R$ (\cref{app:method}).
After a replacement, the head redrafts from the corrected token, and a tie goes to the head.

%% file: sections/04_experiments.tex
\section{Experiments}\label{sec:experiments}

\input{sections/fig_results}
\FloatFigMain
\begin{table*}[t]
\centering
\scriptsize
\setlength{\tabcolsep}{5.2pt}
\renewcommand{\arraystretch}{0.93}
\input{tables/tab_main}
\caption{Speedup ($\times$) over AR in each method's engine, mean $\pm$ standard deviation of three repeats (\cref{app:protocol}). Parentheses: ratio of speedups over LoopSpec. Bold: fastest per model and column. --: not run, as in LoopSpec. \method: no training. BBH: heads trained without BBH questions.}
\label{tab:main}
\end{table*}%

\subsection{Setup}\label{sec:setup}

\paragraph{Models and tasks.}
Following LoopSpec, we evaluate Ouro-1.4B and Ouro-2.6B \citep{zhu2025ouro} with 4 loops, and the Raven-Llama, Raven-OLMo and Raven-TinyLlama checkpoints of \citet{mcleish2025retrofitted} with 32 loops.
The tasks are GSM8K \citep{cobbe2021gsm8k}, MATH500 \citep{lightman2024verify}, BBH \citep{suzgun2023bbh}, HumanEval+ (HE+) and MBPP+ from EvalPlus \citep{liu2023evalplus}, all with LoopSpec's prompts and stop strings.
As in LoopSpec, Raven-TinyLlama skips MATH500 and BBH (\cref{app:data}).
Every model decodes greedily, and Ouro-1.4B and Raven-Llama also sample at \Tone with top-p 0.9 \citep{holtzman2020curious}, as in LoopAhead.

\paragraph{Settings.}
\method drafts up to $G=8$ tokens per step. Ouro advances and checks every loop. Raven advances $d=2$ loops per step under greedy decoding and $d=4$ under sampling, and checks at every doubling of the loop count up to full depth (\cref{tab:app-settings}).
One setting serves every task, chosen on 768 configuration prompts from GSM8K, MATH and MBPP that no evaluation uses (\cref{app:method}).
The optional draft head is one decoder layer with 16.8M parameters, trained on each model's own continuations of prompts that exclude every evaluation question and, for BBH, every BBH question (\cref{app:data}).

\paragraph{Baselines.}
LoopSpec \citep{cho2026loopspec} runs from its official code on SGLang \citep{zheng2024sglang}, and the concurrent CopyLoop \citep{du2026copyloop} from its public server code under greedy decoding, the setting it is evaluated in.
Each baseline runs a configuration chosen for speed on the evaluated prompts themselves, while \method runs its one setting on every task.
LoopAhead has no public code, so we compare with the loop counts it reports (\cref{app:baselines}).

\paragraph{Metric.}
As in concurrent work \citep{liu2026loopahead,du2026copyloop}, a method's speedup is its decode throughput divided by that of AR in its own engine on the same prompts, our AR for \method and SGLang's AR for the baselines.
We compare methods by the ratio of their speedups and average by the geometric mean.
Unless stated otherwise, every timed run uses one NVIDIA H200 in BF16 and decodes one request at a time. Speedups count the decode phase, and counting the prefill too, \methodH keeps 0.71--0.90 of its speedup.

\subsection{Main Results}\label{sec:results}

\paragraph{Speed.}
Averaged over the cells of \cref{tab:main} and without training, \method decodes \nFreeARGreedy\X faster than AR under greedy decoding and \nFreeARTone\X faster at \Tone.
Its ratio of speedups over LoopSpec averages \nFreeVsLoopSpec\X, with a 95\% interval of \nFreeBootLo--\nFreeBootHi\X when prompts are resampled (\cref{app:protocol}), and is above 1 for every model, task and decoding mode.
Over CopyLoop under greedy decoding, the ratio averages \nFreeVsCopyLoop\X.
The draft head raises the two ratios to \nVsLoopSpec\X and \nVsCopyLoop\X (\cref{fig:main}a, \cref{tab:main}).

\paragraph{In the same engine.}
To separate the method from the engine, we also run LoopSpec's schedule in our engine, which at the configurations of \cref{tab:main} takes LoopSpec's loops per output token within 3\% (\cref{app:engine}).
Against this schedule at its fastest configuration per cell, \method is \nEngineFreeMin--\nEngineFreeMax\X as fast per model and mode (\cref{fig:main}c).
\methodH is \nEngineMin--\nEngineMax\X as fast, and faster in every cell and repeat.

\paragraph{Intermediate checks.}
The intermediate checks of \method are those between the admission check and full depth (\cref{sec:method-cycle}).
Without them, \method keeps \nLadXNMin--\nLadXNMax{} of its throughput per model and mode, each at its faster stride (\cref{fig:mechanism}a, \cref{tab:engine}).
At the same stride of 2 on Raven-Llama, it keeps \nLadXtwoN{} (\cref{app:ablations}).
A wrong draft token keeps the tokens behind it running until a check replaces it.
Under greedy decoding, the admission check rejects \nAdmRej\% of the draft tokens it checks, counted in each step up to the first mismatch (\cref{app:depth}).
Of the replacements after it, \nInterShare\% happen below full depth under greedy decoding and \nInterShareTone\% at \Tone. Full depth replaces only \nFullRate\% of the decoded tokens.
On Raven-Llama, these replacements come after a mean of \nMeanDepRvl{} loops, so the positions behind a wrong token run only \nWasteRaven\% of the loops between admission and full depth.
On Ouro, with two intermediate check depths, this share is \nWasteOuro\%, so the checks save the most on Raven.
Under greedy decoding, full depth keeps at least \nKeptFirst\% of the tokens that the first intermediate check writes.

\FloatFigMechanism
\subsection{Ablations}\label{sec:ablations}

\paragraph{Components.}
In \cref{fig:mechanism}a, admitting the head chain's draft tokens without the admission check leaves 0.56--0.63 of the throughput (\cref{app:ablations}), as a wrong token is then found a readout later, with every token behind it.
With one position admitted per step ($G=0$), \methodH keeps 0.52--0.68. Without the n-gram chain, it keeps 0.68--0.79, so the two chains complement each other.
Run as prompt lookup decoding \citep{saxena2023prompt} without the pipeline, the n-gram chain alone reaches only 1.64--2.27\X our AR (\cref{app:engine}).

\paragraph{Shared noise.}
At \Tone, drawing the noise below full depth independently, which \cref{thm:lossless} allows, leaves \nIndepToneOuro{} of the throughput on Ouro-1.4B and \nIndepToneRaven{} on Raven-Llama, as the readouts then disagree with full depth more often (\cref{app:ablations}).

\paragraph{Loops and time per loop.}
\method needs 1.56--2.51\X fewer loops per output token than LoopSpec under greedy decoding and 1.34--1.41\X fewer at \Tone (\cref{fig:main}d).
\methodH needs the fewest loops per output token in every setting (\cref{app:loops}).
Relative to its own AR, the time per loop of \method is at most 1.15\X LoopSpec's, and that of \methodH at most \nLoopTimeMax\X, so with one request per pass the loop count largely decides the speed.

\subsection{Robustness}\label{sec:scaling}

\paragraph{Held-out data.}
On four held-out datasets of the main tasks' types, used neither to choose a setting nor to train a head, the ratio over LoopSpec averages \nFreeHeldVsLoopSpec\X for \method and \nHeldVsLoopSpec\X for \methodH (\cref{fig:main}b, \cref{app:heldout}).

\paragraph{Repetition loops.}
Without the prompts on which the greedy output of any decoder falls into a repetition loop, the ratio over LoopSpec still averages \nFreeNoRep\X for \method and \nNoRep\X for \methodH (\cref{tab:app-norep}).

\paragraph{Reasoning model.}
On the reasoning-tuned Ouro-1.4B-Thinking, with LoopSpec's tasks and sampling (\Tone, top-p 0.7), \method is faster than LoopSpec in every cell (\cref{app:thinking}).
On MATH500, we raise LoopSpec's output budget to 32,768 tokens.
The ratio over LoopSpec grows from 1.64\X to 1.92\X.
Over LoopSpec's schedule in our engine, it grows from 1.58\X to 1.79\X.

\paragraph{Batches.}
With $B$ requests per pass, we choose $G$, the draft chains and Raven's stride per batch size.
Then \methodH is faster than AR on every task up to $B=32$ and, averaged over three tasks, faster than LoopSpec's schedule batched in our engine at every batch size (\cref{fig:mechanism}b, \cref{app:batch}).

\paragraph{Another device and a larger model.}
On an RTX 5090, with all settings chosen again on that GPU, the ratio of speedups of \method over LoopSpec is \nGPUFreeMin--\nGPUFreeMax\X per model and mode (\cref{fig:mechanism}c, \cref{app:gpu}).
On Huginn-0125, a 3.5B recurrent-depth model trained from scratch, \method is faster than both baselines on every task and in every repeat, also without repetition loops (\cref{app:huginn}). Its ratio of speedups over LoopSpec's schedule is \nHugFreeLS\X under greedy decoding and \nHugFreeLSNoRep\X without repetition loops.

\subsection{Lossless in Practice}\label{sec:accuracy}

With every operation in FP32, attention included, the outputs of \methodH, with both draft chains, equal our AR's token for token on \nFPPrompts{} prompts from every task and setting of \cref{tab:main} (\cref{app:accuracy}).
In BF16, the rounding of a matrix product depends on its shape \citep{yuan2025nondeterminism}, so a pass over many tokens can resolve a near tie differently from a pass over one.
On Ouro under greedy decoding, the output of \methodH equals our AR's on \nBFOurs\% of prompts, while our AR equals SGLang's AR on only \nBFAR\%.
The two baselines equal SGLang's AR on \nBFLS\% and \nBFCL\%.
Paired per question with our AR, the pooled accuracy difference of \methodH is \nGreedyDiff{} points under greedy decoding and \nToneDiff{} at \Tone, both with 95\% intervals that contain zero.

%% file: sections/fig_results.tex
\newcommand{\FloatFigMain}{%
\begin{figure*}[t]
\centering
\includegraphics[width=\textwidth]{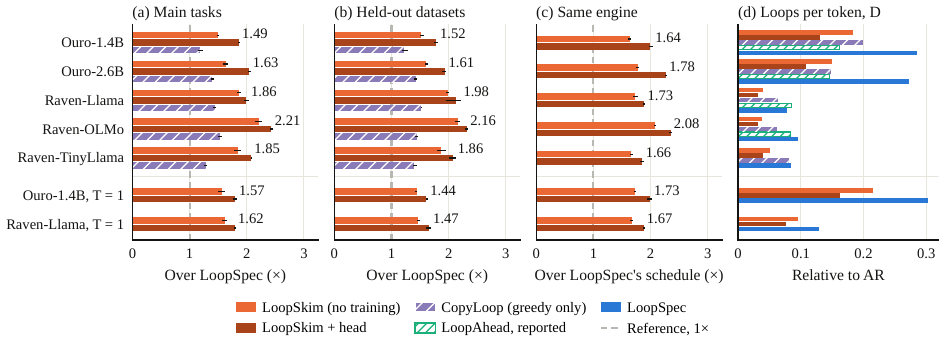}
\caption{Geometric mean over a model's tasks, whiskers: range over three repeats. (a,~b)~Ratio of speedups over LoopSpec on (a)~the main tasks and (b)~four held-out datasets, two on Raven-TinyLlama (\cref{app:heldout}). (c)~Throughput over LoopSpec's schedule run in our engine (\cref{app:engine}).
Numbers in (a--c): \method. (d)~Loops per output token relative to AR ($D$). LoopAhead: as reported in its Figure~3a (\cref{app:loops}).}
\label{fig:main}
\end{figure*}%
}

\newcommand{\FloatTabMain}{%
\begin{table*}[t]
\centering
\scriptsize
\setlength{\tabcolsep}{5.2pt}
\renewcommand{\arraystretch}{0.93}
\input{tables/tab_main}
\caption{Speedup ($\times$) over AR in each method's engine, mean $\pm$ standard deviation of three repeats (\cref{app:protocol}). Parentheses: ratio of speedups over LoopSpec. Bold: fastest per model and column. --: not run, as in LoopSpec. \method: no training. BBH: heads trained without BBH questions.}
\label{tab:main}
\end{table*}%
}

\newcommand{\FloatFigMechanism}{%
\begin{figure*}[t]
\centering
\includegraphics[width=\textwidth]{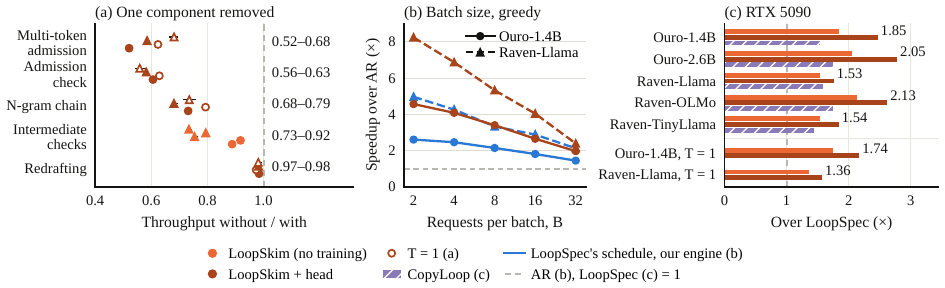}
\caption{(a)~Throughput without one component relative to with it, geometric mean over the five main tasks. Intermediate checks: greedy, five models (\cref{tab:engine}). Others: Ouro-1.4B and Raven-Llama. Circles: Ouro. Triangles: Raven.
(b)~Speedup of \methodH over our AR, geometric mean over GSM8K, MATH500 and MBPP+ (\cref{app:batch}).
(c)~Ratio of speedups over LoopSpec (\cref{app:gpu}). Numbers: \method. Whiskers: range over three repeats.}
\label{fig:mechanism}
\end{figure*}%
}

%% file: tables/tab_main.tex
\begin{tabular}{@{}ll cccccc@{}}
\toprule
Model & Method & GSM8K & MATH500 & BBH & HumanEval+ & MBPP+ & Geo.\ mean \\
\midrule
\multicolumn{8}{c}{\textit{Greedy decoding}} \\
\midrule
 & LoopSpec & 2.88{\tiny$\pm$0.07} & 3.13{\tiny$\pm$0.06} & 2.78{\tiny$\pm$0.06} & 3.19{\tiny$\pm$0.15} & 2.87{\tiny$\pm$0.02} & 2.97{\tiny$\pm$0.03} \\
 & CopyLoop & 3.39{\tiny$\pm$0.20} {\scriptsize(1.18)} & 3.31{\tiny$\pm$0.11} {\scriptsize(1.06)} & 3.23{\tiny$\pm$0.12} {\scriptsize(1.16)} & 5.49{\tiny$\pm$0.18} {\scriptsize(1.72)} & 2.74{\tiny$\pm$0.10} {\scriptsize(0.95)} & 3.53{\tiny$\pm$0.08} {\scriptsize(1.19)} \\
 & \cellcolor{ours}\method & \cellcolor{ours}3.81{\tiny$\pm$0.02} {\scriptsize(1.32)} & \cellcolor{ours}3.79{\tiny$\pm$0.14} {\scriptsize(1.21)} & \cellcolor{ours}\textbf{5.44}{\tiny$\pm$0.17} {\scriptsize(1.96)} & \cellcolor{ours}6.90{\tiny$\pm$0.22} {\scriptsize(2.16)} & \cellcolor{ours}3.18{\tiny$\pm$0.02} {\scriptsize(1.11)} & \multicolumn{1}{>{\columncolor{ours}[\tabcolsep][0pt]}c@{}}{4.44{\tiny$\pm$0.06} {\scriptsize(1.49)}} \\
\multirow{-4}{*}{\makecell[l]{Ouro-\\1.4B}} & \cellcolor{ours}\enspace + head & \cellcolor{ours}\textbf{5.01}{\tiny$\pm$0.02} {\scriptsize(1.74)} & \cellcolor{ours}\textbf{5.24}{\tiny$\pm$0.20} {\scriptsize(1.67)} & \cellcolor{ours}5.20{\tiny$\pm$0.13} {\scriptsize(1.87)} & \cellcolor{ours}\textbf{8.57}{\tiny$\pm$0.16} {\scriptsize(2.69)} & \cellcolor{ours}\textbf{4.38}{\tiny$\pm$0.12} {\scriptsize(1.53)} & \multicolumn{1}{>{\columncolor{ours}[\tabcolsep][0pt]}c@{}}{\textbf{5.52}{\tiny$\pm$0.08} {\scriptsize(1.86)}} \\
\midrule
 & LoopSpec & 3.27{\tiny$\pm$0.04} & 3.39{\tiny$\pm$0.04} & 2.94{\tiny$\pm$0.11} & 3.23{\tiny$\pm$0.04} & 3.17{\tiny$\pm$0.14} & 3.20{\tiny$\pm$0.04} \\
 & CopyLoop & 4.27{\tiny$\pm$0.02} {\scriptsize(1.31)} & 4.90{\tiny$\pm$0.02} {\scriptsize(1.45)} & 4.15{\tiny$\pm$0.09} {\scriptsize(1.41)} & 6.26{\tiny$\pm$0.02} {\scriptsize(1.94)} & 3.33{\tiny$\pm$0.06} {\scriptsize(1.05)} & 4.48{\tiny$\pm$0.02} {\scriptsize(1.40)} \\
 & \cellcolor{ours}\method & \cellcolor{ours}4.36{\tiny$\pm$0.26} {\scriptsize(1.33)} & \cellcolor{ours}5.97{\tiny$\pm$0.24} {\scriptsize(1.76)} & \cellcolor{ours}\textbf{6.40}{\tiny$\pm$0.15} {\scriptsize(2.18)} & \cellcolor{ours}7.10{\tiny$\pm$0.02} {\scriptsize(2.20)} & \cellcolor{ours}3.23{\tiny$\pm$0.07} {\scriptsize(1.02)} & \multicolumn{1}{>{\columncolor{ours}[\tabcolsep][0pt]}c@{}}{5.20{\tiny$\pm$0.11} {\scriptsize(1.63)}} \\
\multirow{-4}{*}{\makecell[l]{Ouro-\\2.6B}} & \cellcolor{ours}\enspace + head & \cellcolor{ours}\textbf{6.14}{\tiny$\pm$0.02} {\scriptsize(1.88)} & \cellcolor{ours}\textbf{7.70}{\tiny$\pm$0.04} {\scriptsize(2.27)} & \cellcolor{ours}6.11{\tiny$\pm$0.16} {\scriptsize(2.08)} & \cellcolor{ours}\textbf{8.77}{\tiny$\pm$0.05} {\scriptsize(2.72)} & \cellcolor{ours}\textbf{4.72}{\tiny$\pm$0.07} {\scriptsize(1.49)} & \multicolumn{1}{>{\columncolor{ours}[\tabcolsep][0pt]}c@{}}{\textbf{6.54}{\tiny$\pm$0.05} {\scriptsize(2.04)}} \\
\midrule
 & LoopSpec & 4.29{\tiny$\pm$0.08} & 6.49{\tiny$\pm$0.12} & 5.60{\tiny$\pm$0.25} & 6.33{\tiny$\pm$0.14} & 5.36{\tiny$\pm$0.11} & 5.56{\tiny$\pm$0.06} \\
 & CopyLoop & 5.31{\tiny$\pm$0.17} {\scriptsize(1.24)} & 9.97{\tiny$\pm$0.23} {\scriptsize(1.54)} & 8.63{\tiny$\pm$0.20} {\scriptsize(1.54)} & 10.94{\tiny$\pm$0.48} {\scriptsize(1.73)} & 6.49{\tiny$\pm$0.08} {\scriptsize(1.21)} & 7.98{\tiny$\pm$0.11} {\scriptsize(1.44)} \\
 & \cellcolor{ours}\method & \cellcolor{ours}6.10{\tiny$\pm$0.37} {\scriptsize(1.42)} & \cellcolor{ours}13.97{\tiny$\pm$0.31} {\scriptsize(2.15)} & \cellcolor{ours}\textbf{11.45}{\tiny$\pm$0.29} {\scriptsize(2.04)} & \cellcolor{ours}14.01{\tiny$\pm$0.41} {\scriptsize(2.21)} & \cellcolor{ours}8.56{\tiny$\pm$0.27} {\scriptsize(1.60)} & \multicolumn{1}{>{\columncolor{ours}[\tabcolsep][0pt]}c@{}}{10.32{\tiny$\pm$0.09} {\scriptsize(1.86)}} \\
\multirow{-4}{*}{\makecell[l]{Raven-\\Llama}} & \cellcolor{ours}\enspace + head & \cellcolor{ours}\textbf{6.71}{\tiny$\pm$0.09} {\scriptsize(1.56)} & \cellcolor{ours}\textbf{15.52}{\tiny$\pm$0.64} {\scriptsize(2.39)} & \cellcolor{ours}10.53{\tiny$\pm$0.29} {\scriptsize(1.88)} & \cellcolor{ours}\textbf{16.10}{\tiny$\pm$0.48} {\scriptsize(2.54)} & \cellcolor{ours}\textbf{9.48}{\tiny$\pm$0.09} {\scriptsize(1.77)} & \multicolumn{1}{>{\columncolor{ours}[\tabcolsep][0pt]}c@{}}{\textbf{11.08}{\tiny$\pm$0.12} {\scriptsize(1.99)}} \\
\midrule
 & LoopSpec & 5.27{\tiny$\pm$0.03} & 4.36{\tiny$\pm$0.07} & 4.57{\tiny$\pm$0.05} & 6.15{\tiny$\pm$0.08} & 6.19{\tiny$\pm$0.01} & 5.25{\tiny$\pm$0.01} \\
 & CopyLoop & 6.67{\tiny$\pm$0.24} {\scriptsize(1.27)} & 6.51{\tiny$\pm$0.08} {\scriptsize(1.49)} & 7.04{\tiny$\pm$0.14} {\scriptsize(1.54)} & 10.24{\tiny$\pm$0.15} {\scriptsize(1.67)} & 10.78{\tiny$\pm$0.42} {\scriptsize(1.74)} & 8.05{\tiny$\pm$0.16} {\scriptsize(1.53)} \\
 & \cellcolor{ours}\method & \cellcolor{ours}8.70{\tiny$\pm$0.48} {\scriptsize(1.65)} & \cellcolor{ours}10.29{\tiny$\pm$0.29} {\scriptsize(2.36)} & \cellcolor{ours}\textbf{11.38}{\tiny$\pm$0.49} {\scriptsize(2.49)} & \cellcolor{ours}14.61{\tiny$\pm$0.21} {\scriptsize(2.38)} & \cellcolor{ours}14.18{\tiny$\pm$0.53} {\scriptsize(2.29)} & \multicolumn{1}{>{\columncolor{ours}[\tabcolsep][0pt]}c@{}}{11.61{\tiny$\pm$0.33} {\scriptsize(2.21)}} \\
\multirow{-4}{*}{\makecell[l]{Raven-\\OLMo}} & \cellcolor{ours}\enspace + head & \cellcolor{ours}\textbf{10.34}{\tiny$\pm$0.04} {\scriptsize(1.96)} & \cellcolor{ours}\textbf{11.39}{\tiny$\pm$0.06} {\scriptsize(2.61)} & \cellcolor{ours}10.82{\tiny$\pm$0.26} {\scriptsize(2.37)} & \cellcolor{ours}\textbf{16.15}{\tiny$\pm$0.28} {\scriptsize(2.63)} & \cellcolor{ours}\textbf{16.55}{\tiny$\pm$0.54} {\scriptsize(2.67)} & \multicolumn{1}{>{\columncolor{ours}[\tabcolsep][0pt]}c@{}}{\textbf{12.78}{\tiny$\pm$0.07} {\scriptsize(2.43)}} \\
\midrule
 & LoopSpec & 5.32{\tiny$\pm$0.20} & -- & -- & 6.69{\tiny$\pm$0.14} & 5.93{\tiny$\pm$0.18} & 5.95{\tiny$\pm$0.06} \\
 & CopyLoop & 6.07{\tiny$\pm$0.15} {\scriptsize(1.14)} & -- & -- & 10.55{\tiny$\pm$0.32} {\scriptsize(1.58)} & 6.88{\tiny$\pm$0.33} {\scriptsize(1.16)} & 7.61{\tiny$\pm$0.01} {\scriptsize(1.28)} \\
 & \cellcolor{ours}\method & \cellcolor{ours}8.50{\tiny$\pm$0.43} {\scriptsize(1.60)} & \cellcolor{ours}-- & \cellcolor{ours}-- & \cellcolor{ours}15.97{\tiny$\pm$0.35} {\scriptsize(2.39)} & \cellcolor{ours}9.81{\tiny$\pm$0.15} {\scriptsize(1.65)} & \multicolumn{1}{>{\columncolor{ours}[\tabcolsep][0pt]}c@{}}{11.00{\tiny$\pm$0.21} {\scriptsize(1.85)}} \\
\multirow{-4}{*}{\makecell[l]{Raven-\\TinyLlama}} & \cellcolor{ours}\enspace + head & \cellcolor{ours}\textbf{9.79}{\tiny$\pm$0.18} {\scriptsize(1.84)} & \cellcolor{ours}-- & \cellcolor{ours}-- & \cellcolor{ours}\textbf{16.98}{\tiny$\pm$0.14} {\scriptsize(2.54)} & \cellcolor{ours}\textbf{11.46}{\tiny$\pm$0.11} {\scriptsize(1.93)} & \multicolumn{1}{>{\columncolor{ours}[\tabcolsep][0pt]}c@{}}{\textbf{12.40}{\tiny$\pm$0.12} {\scriptsize(2.08)}} \\
\midrule
\multicolumn{8}{c}{\textit{Sampling, \Tone, top-p 0.9}} \\
\midrule
 & LoopSpec & 2.34{\tiny$\pm$0.01} & 2.39{\tiny$\pm$0.01} & 2.24{\tiny$\pm$0.01} & 2.59{\tiny$\pm$0.06} & 2.43{\tiny$\pm$0.08} & 2.40{\tiny$\pm$0.02} \\
 & \cellcolor{ours}\method & \cellcolor{ours}3.06{\tiny$\pm$0.04} {\scriptsize(1.31)} & \cellcolor{ours}3.47{\tiny$\pm$0.17} {\scriptsize(1.45)} & \cellcolor{ours}\textbf{4.50}{\tiny$\pm$0.08} {\scriptsize(2.01)} & \cellcolor{ours}5.16{\tiny$\pm$0.07} {\scriptsize(1.99)} & \cellcolor{ours}3.03{\tiny$\pm$0.10} {\scriptsize(1.25)} & \multicolumn{1}{>{\columncolor{ours}[\tabcolsep][0pt]}c@{}}{3.76{\tiny$\pm$0.09} {\scriptsize(1.57)}} \\
\multirow{-3}{*}{\makecell[l]{Ouro-\\1.4B}} & \cellcolor{ours}\enspace + head & \cellcolor{ours}\textbf{3.65}{\tiny$\pm$0.01} {\scriptsize(1.56)} & \cellcolor{ours}\textbf{4.34}{\tiny$\pm$0.00} {\scriptsize(1.82)} & \cellcolor{ours}4.28{\tiny$\pm$0.01} {\scriptsize(1.91)} & \cellcolor{ours}\textbf{5.77}{\tiny$\pm$0.06} {\scriptsize(2.23)} & \cellcolor{ours}\textbf{3.73}{\tiny$\pm$0.07} {\scriptsize(1.53)} & \multicolumn{1}{>{\columncolor{ours}[\tabcolsep][0pt]}c@{}}{\textbf{4.29}{\tiny$\pm$0.01} {\scriptsize(1.79)}} \\
\midrule
 & LoopSpec & 3.53{\tiny$\pm$0.06} & 4.11{\tiny$\pm$0.02} & 4.11{\tiny$\pm$0.01} & 4.00{\tiny$\pm$0.07} & 3.79{\tiny$\pm$0.03} & 3.90{\tiny$\pm$0.01} \\
 & \cellcolor{ours}\method & \cellcolor{ours}4.90{\tiny$\pm$0.14} {\scriptsize(1.39)} & \cellcolor{ours}6.18{\tiny$\pm$0.27} {\scriptsize(1.50)} & \cellcolor{ours}\textbf{8.38}{\tiny$\pm$0.39} {\scriptsize(2.04)} & \cellcolor{ours}6.74{\tiny$\pm$0.01} {\scriptsize(1.69)} & \cellcolor{ours}5.82{\tiny$\pm$0.03} {\scriptsize(1.54)} & \multicolumn{1}{>{\columncolor{ours}[\tabcolsep][0pt]}c@{}}{6.30{\tiny$\pm$0.16} {\scriptsize(1.62)}} \\
\multirow{-3}{*}{\makecell[l]{Raven-\\Llama}} & \cellcolor{ours}\enspace + head & \cellcolor{ours}\textbf{5.62}{\tiny$\pm$0.02} {\scriptsize(1.59)} & \cellcolor{ours}\textbf{7.25}{\tiny$\pm$0.04} {\scriptsize(1.76)} & \cellcolor{ours}8.05{\tiny$\pm$0.07} {\scriptsize(1.96)} & \cellcolor{ours}\textbf{7.58}{\tiny$\pm$0.01} {\scriptsize(1.90)} & \cellcolor{ours}\textbf{6.78}{\tiny$\pm$0.02} {\scriptsize(1.79)} & \multicolumn{1}{>{\columncolor{ours}[\tabcolsep][0pt]}c@{}}{\textbf{7.00}{\tiny$\pm$0.02} {\scriptsize(1.79)}} \\
\bottomrule
\end{tabular}

%% file: sections/05_conclusion.tex
\section{Conclusion}\label{sec:conclusion}

We presented multi-depth speculation, in which the depth of a looped LM's pipeline is the verifier. Every check depth compares the readouts in flight with the draft tokens after them and replaces the first one it contradicts, without an extra pass.
Fixed random draws keep \method lossless under sampling for any number of checks and draft chains.
With draft tokens copied from the context and no training, \method is faster than LoopSpec on every model and task we evaluate on the H200. Its checks between admission and full depth speed it up on every model, and an optional draft head adds further speed.

%% file: sections/06_limitations.tex
\section*{Limitations}

We evaluate looped LMs of up to 3.5B parameters on math, code and reasoning tasks. We run them on H200 and RTX 5090 GPUs with up to 32 requests per batch.
\method trades compute for latency. Per output token, \method and \methodH run 4.74--7.74\X as many loops as AR over all token positions, which may raise the energy per token and leaves less compute for other requests. Its settings therefore depend on the batch size, and we choose them per batch size (\cref{sec:scaling}). Integrating it into a production serving engine is future work.
The margin over LoopSpec narrows at long contexts (\cref{app:batch}).
The n-gram chain gains most where an output repeats text from the context or from itself (\cref{sec:results}).

The optional draft head needs 2.24--4.16 H200-hours to decode the model's own outputs and minutes of training per model (\cref{app:method}).
On BBH, whose questions its training data exclude, \methodH is slightly slower than \method in every cell.

%% file: sections/app_A_setting.tex
\renewcommand{\topfraction}{0.9}\renewcommand{\dbltopfraction}{0.9}\renewcommand{\textfraction}{0.08}
\renewcommand{\floatpagefraction}{0.85}\renewcommand{\dblfloatpagefraction}{0.85}\setcounter{dbltopnumber}{3}
\noindent\textbf{Guide to the appendices.}
\Cref{app:setting} gives the setting, \cref{app:proof} the proof of \cref{thm:lossless}, and \cref{app:results} the results behind \cref{sec:experiments}.

\section{Experimental Setting}\label{app:setting}

\subsection{Models and Data}\label{app:data}

\paragraph{Models.}
Ouro-1.4B and Ouro-2.6B \citep{zhu2025ouro}, the base checkpoints rather than the Thinking ones, loop their whole layer stack $R=4$ times.
The Raven checkpoints of \citet{mcleish2025retrofitted}, retrofitted from Llama \citep{grattafiori2024llama3}, OLMo \citep{olmo2025olmo2} and TinyLlama \citep{zhang2024tinyllama}, run a prelude, $R=32$ loops of a recurrent block and a coda, and start the recurrent state of every position from a truncated normal.
A readout below full depth applies the coda to the state at that depth.
Raven-Llama and Raven-OLMo have about 1.4B parameters, and Raven-TinyLlama about 0.8B.
Ouro-1.4B-Thinking, Ouro's reasoning-tuned checkpoint, runs with its chat template and thinking mode enabled. No draft head was trained for it: \method runs on it without training, and \methodH uses the Ouro-1.4B head.

\paragraph{Tasks.}
\Cref{tab:app-data} lists every evaluated dataset.
The five main tasks use the prompts, few-shot formats and stop strings of LoopSpec's client on lm-eval 0.4.12 \citep{biderman2024lessons}, and as in LoopSpec, Raven-TinyLlama runs three of them, since the prompts of MATH500 and BBH with their output limits exceed its 2,048-token context.

\paragraph{Draft-head data.}
One pool of 9,025 prompts, with the source counts of LoopAhead's drafter pool \citep{liu2026loopahead}, trains the draft heads of all five models.
It holds 2,000 prompts each from the training sets of GSM8K and MATH \citep{hendrycks2021math} and from the Python problems of CodeAlpaca, 2,025 BBH questions and 1,000 UltraChat dialogues \citep{ding2023ultrachat}.
Its 75 BBH questions per subtask come after the first 20, which are the evaluated ones.
The pool drops near duplicates of every evaluation question and few-shot demonstration of the main tasks, and every prompt that holds half or more of a question's distinctive word 13-grams.
BBH has no official training split, so on BBH the main table and the runs derived from it use draft heads retrained by the same recipe without any BBH question.

\begin{table*}[t]
\centering
\scriptsize
\setlength{\tabcolsep}{3pt}
\begin{tabular}{@{}l r r l l@{}}
\toprule
Data & Prompts & Output limit & Subset & Prompt format \\
\midrule
GSM8K \citep{cobbe2021gsm8k} & 1,319 & 512 & test set & LoopSpec's \\
MATH500 \citep{hendrycks2021math,lightman2024verify} & 500 & 2,048 & all & LoopSpec's \\
BBH \citep{suzgun2023bbh} & 540 & 1,024 & first 20 per subtask & LoopSpec's \\
HE+ \citep{liu2023evalplus,chen2021humaneval} & 164 & 1,024 & all & LoopSpec's \\
MBPP+ \citep{liu2023evalplus,austin2021mbpp} & 378 & 1,024 & all & LoopSpec's \\
\midrule
SVAMP \citep{patel2021svamp} & 1,000 & 512 & all, train and test & as GSM8K \\
OlympiadBench \citep{he2024olympiadbench} & 199 & 2,048 & 200 of 674 English text-only math problems & as MATH500 \\
MMLU-Pro \citep{wang2024mmlupro} & 278 & 1,024 & 20 test questions per category & its 5-shot chain of thought \\
BigCodeBench \citep{zhuo2025bigcodebench} & 200 & 1,024 & 200 of 1,140 Complete prompts (v0.1.4) & as HE+ \\
\midrule
Spec-Bench \citep{xia2024unlocking} & 480 & 1,024 & all, 80 per category & Vicuna template, first turn \\
PG-19 \citep{rae2020compressive} & 32 & 256 & 32 books, contexts of 1K--32K tokens & book text \\
\midrule
\multicolumn{5}{c}{\textit{Ouro-1.4B-Thinking (LoopSpec's thinking tasks, thinking mode on)}} \\
\midrule
GSM8K & 1,319 & 4,096 & test set & chat template \\
MATH500 & 500 & 8,192 / 32,768 & all & chat template \\
AIME24, AIME25 & 30 each & 16,384 / 32,768 & all & chat template \\
\bottomrule
\end{tabular}
\caption{Main tasks (top), held-out datasets and other data (middle, \cref{app:heldout,app:batch}) and the thinking tasks (bottom, \cref{app:thinking}). Prompts: count after filtering. Output limit: maximum number of new tokens. Held-out samples are drawn with seed 24, and every prompt stops at the stop strings of its task.}
\label{tab:app-data}
\end{table*}

\subsection{Implementation of \texorpdfstring{\method}{LoopSkim}}\label{app:method}

\begin{algorithm}[t]
  \footnotesize
  \caption{\method decoding. $\hat{x}^{(\ell)}_i$: readout token of depth $\ell$ at position $i$, \cref{eq:shared-noise} ($g \equiv 0$ under greedy decoding). $\mathcal{B}$: active groups, each a run $[P_0, P_0{+}n)$ of positions at one depth.}
  \label{alg:cycle}
  \begin{algorithmic}[1]
    \Require prompt $x_{1:L}$; loops $R$; stride $d$; check depths $\Lambda = \{\ell_1{=}d < \ell_2 < \dots < \ell_{|\Lambda|}{=}R\}$; draft chains $k = 1, \dots, C$ (1: n-gram chain, 2: optional draft head); draft length $G$
    \Ensure committed tokens $x_{L+1:c}$
    \State $x_{L+1} \gets \textsc{Prefill}(x_{1:L})$; \ $P, c \gets L{+}1$; \ $\mathcal{B} \gets \emptyset$
    \While{not $\textsc{Stop}(x_{\le c})$}
      \State $u^k_{1:G} \gets \textsc{Draft}_k(x_{\le P}, h^{(\ell_1)}_{<P})$ for $k = 1, \dots, C$
      \State $\hat{x} \gets \textsc{Pass}\bigl(\{(x_P, u^k_{1:G})\}_{k} \text{ at } \ell_1;\ \mathcal{B} \text{ by } d\bigr)$
      \For{$\ell = \ell_{|\Lambda|}, \dots, \ell_2$} \Comment{deepest first}
        \State $\beta = [P_0, P_0{+}n) \gets$ group of $\mathcal{B}$ at depth $\ell$, if any
        \State $m \gets \min\{1 \le j \le n : x_{P_0+j} \ne \hat{x}^{(\ell)}_{P_0+j}\}$ \Comment{\eqref{eq:trunc}}
        \If{$m$ exists} \Comment{replace}
          \State $P \gets P_0{+}m$; \ $x_P \gets \hat{x}^{(\ell)}_{P}$
          \State \textbf{if} $\ell = R$ \textbf{then} $c \gets P$; $\mathcal{B} \gets \emptyset$
          \State \textbf{else} drop the positions $\ge P$ from $\mathcal{B}$
          \State \textbf{next step}
        \ElsIf{$\ell = R$} \Comment{commit}
          \State $c \gets P_0{+}n$; \ $\mathcal{B} \gets \mathcal{B} \setminus \{\beta\}$
        \EndIf
      \EndFor
      \State $a_k \gets \max\{j \le G : u^k_{1:j} = \hat{x}^{(\ell_1,k)}_{P+1:P+j}\}$ \Comment{\eqref{eq:admit}}
      \State $k^\star \gets \max \arg\max_k a_k$; \ $a \gets a_{k^\star}$ \Comment{head on ties}
      \State $x_{P+1:P+a} \gets u^{k^\star}_{1:a}$; \ $\mathcal{B} \gets \mathcal{B} \cup \{[P, P{+}a{+}1)\}$
      \State $P \gets P{+}a{+}1$; \ $x_P \gets \hat{x}^{(\ell_1,k^\star)}_{P}$
    \EndWhile
    \State \Return $x_{L+1:c}$ up to its first stop
  \end{algorithmic}
\end{algorithm}

\begin{table}[t]
  \centering
  \footnotesize
  \setlength{\tabcolsep}{3pt}
  \begin{tabular}{@{}>{\raggedright\arraybackslash}p{2.6cm}
                     >{\raggedright\arraybackslash}p{1.3cm}
                     >{\raggedright\arraybackslash}p{3.0cm}@{}}
    \toprule
    Setting & Ouro & Raven \\
    \midrule
    Draft tokens per chain $G$ & 8 & 8 \\
    Stride $d$ & 1 & greedy: 2, $T>0$: 4 \\
    Check depths $\Lambda$ & 1, 2, 3, 4 & greedy: 2, 4, 8, 16, 32 \newline $T>0$: 4, 8, 16, 32 \\
    Tokens per pass & 36 (45) & greedy: 144 (153) \newline $T>0$: 72 (81) \\
    Head draws, $T>0$ & at $0.7\times T$ & at $T$ \\
    Pruning margin, $T>0$ & 3 nats & -- \\
    \bottomrule
  \end{tabular}
  \caption{Default settings. Tokens per pass: $(C+R/d-1)(1+G)$, with the draft head in parentheses. Pruning margin: readouts below full depth drop tokens more than 3 nats below their top token.}
  \label{tab:app-settings}
\end{table}

\Cref{alg:cycle} gives the decoding loop, with the order of events of \cref{sec:method-cycle}, and \cref{tab:app-settings} the default settings.
\method runs the n-gram chain alone ($C=1$), and \methodH adds the draft head ($C=2$). All other settings are shared.

\paragraph{N-gram chain.}
The host indexes every $n$-gram of the prompt and the generated text, for $n$ from 2 to 4, by the position where it last ended.

\paragraph{Draft head.}
The head reads the last $w=32$ positions. At each, a linear map takes the depth-$\ell_1$ state $h^{(\ell_1)}_i$ (on Raven, the normalized coda output at depth $d$) concatenated with the embedding of $x_{i+1}$, and $G-1$ learned query vectors follow the window.
One causal decoder layer of width 1,024, with 8 attention heads and an MLP width of 2,048, and a linear map to the model's hidden size give $\tilde{h}_j$, the predicted depth-$\ell_1$ state at position $P{+}j{-}1$.
The model's final normalization and output head turn $\tilde{h}_j$ into $q_{\phi,j}$, and the draft token $u_j$ is its readout token within the 8,192 most frequent depth-$\ell_1$ target tokens, the size of LoopAhead's draft vocabulary.
As its draft tokens first meet the admission check, the head is trained toward $p_{\ell_1}$ on the model's own continuations:
\begin{equation}
\begin{split}
  \mathcal{L}(\phi) = {} & \frac{1}{G} \sum_{j=1}^{G} \Bigl[ \mathrm{CE}\bigl( p_{\ell_1}(\cdot \mid x_{<P+j}),\, q_{\phi,j} \bigr) \\
  & \quad + \lambda \, \mathrm{SmoothL1}\bigl( \tilde{h}_j,\, h^{(\ell_1)}_{P+j-1} \bigr) \Bigr],
\end{split}
  \label{eq:headloss}
\end{equation}
with $\lambda = 0.2$, for 16,000 steps of batch size 256 at a learning rate of $3\times10^{-4}$ with cosine decay.
The head has 16.8M parameters. On Raven, one head trained on depth-2 states serves both strides. On Raven-Llama, it raises the throughput of \method by 1.11\X under sampling and by 1.07\X under greedy decoding (raw throughput in \cref{tab:engine}).
Decoding the model's own greedy continuations took 2.24--4.16 H200-hours per model, and training a head takes 4.2--16.9 minutes on one H200.

\paragraph{Sampler.}
The Gumbel noise $g_{s,i,v}$ of \cref{eq:shared-noise} comes from a counter-based generator \citep{salmon2011parallel} whose inputs are the seed, the position and the token.
The top-$p$ set keeps the most probable tokens until their mass reaches $p$, plus every token tied with the last one, as LoopSpec's sampler does.
As \cref{thm:lossless} uses \cref{eq:shared-noise} only at full depth, readouts below full depth and the draft head may draw differently, as the pruning margin and head temperature of \cref{tab:app-settings} do.
We computed exactly, from AR's readout distributions at \Tone on its own outputs for the first 20 prompts of each main task, how often a readout below full depth draws the same token as the full-depth readout at its position.
With this noise, on Ouro-1.4B and Raven-Llama, the probability is 99.4--99.9\% of $1-\mathrm{TV}$, the largest that any coupling of the two distributions attains \citep{daliri2024coupling}. With independent draws, it is 79.6--94.8\%.
Raven's initial states $\xi_{s,i}$ come from the same kind of generator, with seed, position and hidden dimension as inputs and the seed offset by $2^{40}$, mapped by the inverse CDF to the released truncated normal.

\paragraph{Choice of settings.}
One setting serves every task. We chose it on 768 configuration prompts, 256 each from the training sets of GSM8K and MATH and from the training and validation sets of MBPP \citep{austin2021mbpp} without MBPP+ problems, in the formats of GSM8K, MATH500 and MBPP+.
Changing one setting at a time on these prompts, throughput stays within 3\% of the values we use for $G \in \{2, 4, 8, 16\}$, 4,000 to 32,000 training steps of the head and one step of the head temperature or the pruning margin, and no value tested is more than 1\% faster.
The check depths and the stride matter more. Checking Ouro only at depths 1 and 4 gives 0.88 of the throughput, and on Raven-Llama under greedy decoding, a stride of 4 gives 0.87 and other check depths give 0.56--0.96.

\subsection{Baselines}\label{app:baselines}

\input{tables/tab_design}

\paragraph{Entries of Table~\ref{tab:design}.}
LoopSpec's gate at its second proposal depth only opens a fallback branch, so LoopSpec, like WaveFront and DAS-Wave, stops a draft token only at full depth.
CopyLoop keeps the longest prefix of its copied tokens that the readouts at its check depth match and replaces the next one by that readout token.
LoopAhead's drafter proposes a tree of candidate tokens, and the drafts of the other decoders form one chain.
The entries follow LoopSpec \citep[Sec.~4.3]{cho2026loopspec}, WaveFront \citep[Alg.~1]{ha2026wavefront}, DAS-Wave \citep[Secs.~3.2--3.4 and 4.1]{li2026das}, LoopAhead \citep[Table~7 and App.~A.5]{liu2026loopahead} and CopyLoop \citep[Secs.~2--3 and App.~B.2]{du2026copyloop}, at the depths of their reported runs and our CopyLoop calibration.

\paragraph{LoopSpec.}
LoopSpec runs from its official code on SGLang \citep{cho2026loopspec,zheng2024sglang} with SGLang's standard linear layers, for both LoopSpec and its AR, since FlashInfer's TinyGEMM kernels (version 0.6.12), which LoopSpec's paper uses on an RTX PRO 6000, hang or return wrong results on the H200 in our tests.
Its candidate proposal depths are 1 and 1:2 on Ouro, and on Raven 2, 4, 2:8, 2:10 and 4:8 under greedy decoding and 2:8 and 4:8 at \Tone.
On Raven, each cell times every candidate once on the full task and keeps the fastest, 2:8 or 2:10 in every greedy cell but Raven-OLMo MATH500 (4:8) and mostly 4:8 at \Tone. On Ouro, 1:2 is faster than 1 on GSM8K and MATH500 of both models under greedy decoding and runs in every cell.
Timed by decode time as in its paper, its speedup over its own AR is 0.94--1.21 times the published one per greedy cell, 1.02 on average.

\paragraph{CopyLoop.}
CopyLoop \citep{du2026copyloop} runs from its public code with its SGLang server, flags and environment, and the same linear layers as LoopSpec. We run it under greedy decoding, the setting its paper evaluates, and its speedup divides by its own run of SGLang's AR.
Each cell runs the configuration fastest in a calibration pass on its own prompts over CopyLoop's grid \citep[Sec.~3.3]{du2026copyloop}, check depths $d_s \in \{1, 2\}$ on Ouro and $\{4, 8, 16\}$ on Raven with copy lengths $K \in \{2, 4, 6, 8\}$ on Ouro and $\{4, 6, 8\}$ on Raven. On Raven we add $d_s=2$ with $K=4$, as larger $K$ exceeds the executor's limit of 80 tokens per block.
It keeps $d_s=1$ on Ouro and $d_s=4$ on Raven. On the PG-19 contexts, the same calibration runs per context length and mostly keeps $d_s=2$ on Ouro and $d_s=8$ on Raven-Llama. A second pass with $K=2$ and $d_s \in \{2, 4, 8, 16\}$ in every Raven cell reached 0.75--0.97 of the kept configuration's throughput.
On the models that both papers run, CopyLoop's ratio of speedups over LoopSpec measured here is 0.94--1.10 times the one in its Table~1.

\paragraph{LoopAhead and other decoders.}
We found no public code for LoopAhead \citep{liu2026loopahead} or DAS-Wave \citep{li2026das}. LoopAhead runs LoopSpec at 1:2 on Ouro and 2:8 on Raven. Its loops per output token do not depend on the GPU or the engine, and \methodH needs fewer than LoopAhead reports on every model it reports (\cref{tab:app-loops}).
WaveFront \citep{ha2026wavefront} commits at most one token per $T_d$ loop calls, so its $D \ge T_d/R$, which is 0.25 with its $T_d=1$ on Ouro-2.6B, above the values of \methodH in \cref{tab:app-loops}.
In CopyLoop's measurements, WaveFront is slower than LoopSpec in 23 of 24 model--task pairs. Prompt lookup decoding \citep{saxena2023prompt}, which verifies copied tokens with full-depth passes, is slower than LoopSpec in every pair \citep[Table~1]{du2026copyloop}.

\subsection{Protocol}\label{app:protocol}

\paragraph{Timing.}
Except in \cref{app:batch}, every speed is the decode throughput, decoded tokens over decode time, of one request at a time on a single H200 in BF16.
Each prompt is decoded to the length of \methodH's output (until its end token or a stop string, else the output limit), and our AR and \method decode exactly that many tokens. The servers stop at their own end token or stop string. Where the output of our AR or \method reaches its own end token or stop string earlier, it decodes on to that length. Without the prompts where this happens, the speedup of \method over our AR is higher in every setting, by 0.5--3.1\%.
\method and our AR share one PyTorch 2.13 engine with a paged KV cache \citep{kwon2023vllm}, the FlashAttention kernels of vLLM 0.32, CUDA graphs and the same sampler, and a step runs as a single CUDA graph. LoopSpec and CopyLoop run on SGLang 0.5.14 with PyTorch 2.11.
Our decode time includes the host's n-gram lookup and synchronizations and excludes the prefill and graph capture. Counting the prefill, \methodH keeps 0.71--0.90 of its decode-only speedup over our AR in \cref{tab:main}.
Each of our runs first decodes 32 untimed tokens of one prompt, and the servers' start-up request is dropped. For every method, a cell's throughput is its decoded tokens over its decode time, both summed over the prompts.

\paragraph{Repeats and intervals.}
Each cell runs three repeats, and tables give their mean. An average over cells is the geometric mean of these means, and whiskers give its range over the three repeats.
Resampling the prompts within each task 2,000 times, with the same draws for every model, the average ratio of speedups over LoopSpec has a 95\% percentile interval of \nFreeBootLo--\nFreeBootHi{} for \method and \nBootLo--\nBootHi{} for \methodH.

\paragraph{Sampling.}
At \Tone the main runs use top-$p$ 0.9, as in LoopAhead, and each repeat averages the seeds 0, 42 and 271828.
As with a fixed seed per request in serving engines, prompts decoded with one seed share the noise at equal positions, and \cref{thm:lossless} also holds with a request identifier among the generator's inputs.
LoopSpec's AR run with the first seed is its denominator for all seeds.
Under greedy decoding, our engine draws Raven's initial states with seed 0, and the SGLang servers draw them from their own seeded generator. On Raven and at \Tone, the baselines therefore decode texts of their own.

\paragraph{Ablations.}
Ablations run three repeats with seed 0.

\paragraph{Accuracy.}
Accuracy follows LoopSpec's scoring flow, replaying the decoded texts through lm-eval and running generated code in a sandbox.
Each arm decodes until its own end token, else the output limit. At \Tone both arms use the same noise, and a question's score is its mean over the three seeds.
The pooled difference sums the per-question differences of all models. Its 95\% interval comes from 10,000 bootstrap draws of questions within each task, with the same draws for every model, and its $p$-value from a two-sided sign-flip test.

%% file: tables/tab_design.tex
\begin{table}[t]
\centering
\footnotesize
\setlength{\tabcolsep}{2.0pt}
\resizebox{\columnwidth}{!}{%
\begin{tabular}{@{}llccccc c@{}}
\toprule
 & & & & \multicolumn{2}{c}{Checks below $R$} & & \\
\cmidrule(lr){5-6}
Method & \makecell[l]{Draft\\source} & \makecell{Draft\\chains} & \makecell{Draft tokens\\/ step} & Ouro & Raven & \makecell{On contra-\\diction} & Sampling \\
\midrule
LoopSpec & readout & 1 & 1 & -- & -- & -- & \cmark \\
WaveFront & readout & 1 & 1 & -- & -- & -- & \xmark \\
DAS-Wave & readout & 1 & $\le$24 & -- & -- & -- & \xmark \\
LoopAhead & drafter$^{*}$ & tree & $\le$3 & 2 & 4, 8$^{\dagger}$ & paused & \cmark \\
CopyLoop & context & 1 & $K$ & 1 & 4 & replaced & \xmark \\
\midrule
\rowcolor{ours}\method & context$^{\ddagger}$ & $C$ & $G$ & 1, 2, 3 & 2, 4, 8, 16$^{\S}$ & replaced & \cmark \\
\bottomrule
\end{tabular}}
\caption{Self-speculative decoders for looped LMs. Checks below $R$: depths, in loops, at which a check can stop a draft token ($R=4$ on Ouro, 32 on Raven). Sampling: evaluated with lossless sampling (\xmark: not evaluated). --: none. $^{*}$Trained. $^{\dagger}$2, 8 on Raven-TinyLlama. $^{\ddagger}$Plus an optional draft head. $^{\S}$4, 8, 16 under sampling.}
\label{tab:design}
\end{table}

%% file: sections/app_B_proof.tex
\section{Proof of Losslessness}\label{app:proof}

We prove \cref{thm:lossless} in exact arithmetic for the pipeline step of \cref{alg:cycle}. The FP32 identity runs (\cref{app:accuracy}) test the code that implements it.

\paragraph{Notation.}
Position $i$ takes $x_i$ as input, and its readout at depth $\ell$ gives $\hat{x}^{(\ell)}_{i+1}$. At the admission check, each draft chain computes its own positions $P+1, \dots, P+G$.
A KV entry belongs to a cache $\kappa$ (Ouro: one per loop. Raven: the prelude, each loop and a coda per readout depth) and a position $i$. Entries sit in fixed-size blocks that each sequence reads through its own block table \citep{kwon2023vllm}.
Let $p_R(\cdot \mid x_{\le i})$ be AR's full-depth readout after the prefix $x_{\le i}$ (on Raven also after $\xi_{s,\le i}$), and let $y_{i+1} = \arg\max_{v \in \mathcal{V}_p} [\log p_R(v \mid x_{\le i}) + g_{s,i+1,v}]$ be the token that \cref{eq:shared-noise} draws from it ($g \equiv 0$ under greedy decoding). AR sets $x_{i+1} = y_{i+1}$.
The host keeps the committed tokens $x_1, \dots, x_c$ followed by the uncommitted ones. An entry or buffered state at $i$ is consistent if it equals what AR computes there from the current tokens $x_{\le i}$, which for a draft chain at the admission check are $x_{\le P}$ and that chain's draft tokens.

\paragraph{Invariants.}
\begin{enumerate}\setlength{\itemsep}{0pt}
  \item[(I1)] Position $i$ reads entries of its own cache and sequence at positions $\le i$ only.
  \item[(I2)] In every layer, a call stores the entries of all its positions before its attention runs.
  \item[(I3)] Groups are formed in position order and advance $d$ loops per step, so in every cache a group runs after all groups at earlier positions.
  \item[(I4)] A check that contradicts the token at $i^\ast$, at any depth, keeps the tokens before $i^\ast$, writes the readout token at $i^\ast$, drops every position $\ge i^\ast$ and restarts there.
  \item[(I5)] Padding tokens write to a scratch block, a second chain writes to private blocks that are moved into the main blocks only if it is kept, and entries that the admission check has not completed are rewritten before any group reads them.
  \item[(I6)] A token is committed only from the full-depth readout at a position whose inputs are all committed. When a group reaches full depth its first input is committed, and each later input becomes committed once the readout before it equals it.
  \item[(I7)] On Raven, $\xi_{s,i}$ is a fixed function of the seed, the position and the hidden dimension.
\end{enumerate}

\begin{lemma}\label{lem:reads}
  Whenever position $i$ reads an entry $(\kappa, i')$ with $i' \le i$, the entry is consistent with $x_{\le i'}$.
\end{lemma}

\begin{proof}
By induction over calls in execution order, from the prefill.
The input of position $i$ in $\kappa$ is its output in the previous cache (or its embedding, a buffered state and on Raven $\xi_{s,i}$), and its entry also depends on the earlier entries of $\kappa$ it reads (I1). These are written in the same call before the read (I2) or in an earlier step (I3).
A new token at $i^\ast$ (I4) can only make entries at positions $\ge i^\ast$ inconsistent. These positions are dropped, and the restarted positions rewrite these entries before any later position reads them (I2, I3). Writes that are not rewritten go to blocks that no kept position reads (I5).
\end{proof}

\begin{proof}[Proof of \cref{thm:lossless}]
Suppose $x_{\le c}$ are committed and equal AR's tokens. By (I6), the token at $c+1$ comes from the full-depth readout at position $c$, whose inputs are committed.
By \cref{lem:reads} and (I7), this readout is $p_R(\cdot \mid x_{\le c})$, and drawn with $g_{s,c+1,\cdot}$ it gives $y_{c+1}$, which is AR's token.
Draft tokens, the choice among chains, checks below full depth and replacements only decide what is computed ahead of time. They never enter a committed value, so readouts below full depth and the head may also draw from other distributions.
At $T>0$, $g$ is i.i.d.\ Gumbel(0,\,1) and depends only on the seed, the position and the token, so by the Gumbel-max trick AR with this noise is ordinary top-$p$ sampling, and \method commits the same sequence.
Both break ties toward the lowest token index and stop by the same rule.
\method also makes progress. The first position of the oldest group reads only committed tokens, so no check drops it, and within $R/d$ steps it commits a token.
\end{proof}

\subsection{The Initial-State Condition on Raven}\label{app:proof-raven}
If each draft chain drew its own initial states and the chain with more admitted tokens were kept, the kept states would be selected for agreeing with the draft tokens. If two chains draft A and each emits A with probability 1/2, keeping the one that emits A gives A with probability 3/4.
With $\xi_{s,i}$ a function of the seed and the position (I7), every chain computes the same function of equal prefixes, and these states follow the released model's initial-state distribution (\cref{app:method}), so AR with them is still a sample of the released model.

%% file: sections/app_C_results.tex
\section{Results}\label{app:results}

\begin{table*}[t]
\centering
\scriptsize
\setlength{\tabcolsep}{2.2pt}
\input{tables/tab_heldout}
\caption{Ratio of speedups, mean of three repeats. Mean: geometric mean over the held-out datasets. Spec-Bench: geometric mean over its six categories. Olymp.: OlympiadBench. BCB: BigCodeBench.}
\label{tab:heldout}
\end{table*}

\begin{table*}[t]
\centering
\scriptsize
\setlength{\tabcolsep}{5.2pt}
\renewcommand{\arraystretch}{0.93}
\input{tables/tab_main_norep}
\caption{As \cref{tab:main}, without the prompts on which the greedy output of any of its decoders falls into a repetition loop in any repeat. The same prompts are left out at \Tone.}
\label{tab:app-norep}
\end{table*}

\begin{table*}[t]
\centering
\scriptsize
\setlength{\tabcolsep}{3.2pt}
\input{tables/tab_engine}
\caption{Raw throughput in \cref{tab:main} (top) and the controls of \cref{app:engine}. Geometric mean over tasks of the per-cell means over three repeats. Ratios in our engine: per repeat, the geometric mean over tasks of the per-cell throughput ratio, then the mean over repeats. $^{*}$Raven, greedy: stride 4 without the checks, 2 with them.}
\label{tab:engine}\label{tab:app-raw}
\end{table*}

\begin{table}[t]
\centering
\scriptsize
\setlength{\tabcolsep}{3pt}
\input{tables/tab_depth}
\caption{Replacements per check depth of \method on the prompts of \cref{app:engine}, pooled over each model's tasks. Rejected: share of the draft tokens checked. Below full depth: share of the later replacements made at an intermediate depth. Rate: full-depth replacements per decoded token. Mean depth: mean check depth, in loops, of the later replacements ($R=4$ on Ouro, 32 on Raven). All: pooled over models. All + head: the same for \methodH. \Tone: seed 0.}
\label{tab:app-depth}
\end{table}

\begin{table*}[t]
\centering
\scriptsize
\setlength{\tabcolsep}{4pt}
\input{tables/tab_loops}
\caption{The factors of \cref{eq:speedup} on the prompts of \cref{tab:main}, pooled over tasks, and the loops over all token positions (\cref{app:loops}). Last block: LoopAhead's Figures~3a and~3b (greedy). LoopSpec's time per loop: its official code over SGLang's AR. Bold: lowest per column in each block but $\tau$.}
\label{tab:app-loops}
\end{table*}

\begin{table*}[t]
\centering
\scriptsize
\input{tables/tab_app_ablation}
\caption{Throughput with one component removed, relative to with it, for \methodH (top) and \method (bottom). Geometric mean over tasks, mean of three repeats (range). \Tone: seed 0. $^{*}$Head chain alone, relative to it with the admission check. $^{\dagger}$Raven, greedy: stride 4 without the checks, 2 with them. $^{\ddagger}$Stride 2 in both arms. $^{\S}$Noise below full depth drawn independently of full depth's.}
\label{tab:app-ablation}
\end{table*}

\begin{table}[t]
\centering
\scriptsize
\setlength{\tabcolsep}{2.6pt}
\input{tables/tab_rscan}
\caption{Raven-Llama run with $R$ loops. Speedup over our AR of \methodH and of LoopSpec's schedule in our engine, and their ratio. Geometric mean over tasks, mean of three repeats.}
\label{tab:app-rscan}
\end{table}

\begin{table*}[t]
\centering
\scriptsize
\setlength{\tabcolsep}{4pt}
\input{tables/tab_acc}
\caption{Accuracy (\%), each decoder run to its end token or the output limit. \Tone: mean over three seeds.}
\label{tab:app-acc}
\end{table*}

\begin{table*}[t]
\centering
\scriptsize
\setlength{\tabcolsep}{4pt}
\input{tables/tab_gpu}
\caption{Speedup ($\times$) over AR in each method's engine on an RTX 5090, 100 prompts per cell, mean of three repeats. Parentheses: ratio of speedups over LoopSpec. Bold: fastest per column.}
\label{tab:app-gpu}
\end{table*}

\subsection{Held-Out Datasets, Spec-Bench and Repetition Loops}\label{app:heldout}

The four held-out datasets of \cref{tab:heldout} are of the main tasks' types and played no part in choosing the settings or training the draft heads.
We remove three questions that share half or more of their distinctive word 13-grams with a draft-head training question or a main-table question, which leaves 1,677 prompts. Raven-TinyLlama, with its 2,048-token context, runs SVAMP and BigCodeBench only.
\method and \methodH are faster than LoopSpec in all 26 cells. Over CopyLoop, \methodH is faster in all 18 greedy cells and \method in 17.

\paragraph{Spec-Bench.}
Spec-Bench \citep{xia2024unlocking} covers multi-turn conversation, translation, summarization, question answering, math word problems from GSM8K and retrieval-augmented generation (\cref{tab:app-data}). No draft head was trained on its questions, and each baseline runs its fastest configuration on the whole set.
Averaged over the categories, \method and \methodH are faster than LoopSpec on every model and mode and than CopyLoop on every model (\cref{tab:heldout}).

\paragraph{Repetition loops.}
Greedy outputs of these models sometimes repeat a passage until the output limit. An output falls into such a repetition loop when more than half of its 32-token windows appeared earlier in it.
\Cref{tab:app-norep} repeats \cref{tab:main} without the prompts on which the greedy output of any of its decoders does so in any repeat: our AR, \method, \methodH, LoopSpec, CopyLoop and the SGLang ARs.
The ratio of speedups over LoopSpec then averages \nFreeNoRep\X for \method and \nNoRep\X for \methodH. Over CopyLoop, it averages 1.27\X and 1.48\X.
\method stays faster than LoopSpec in every cell, and under greedy decoding it decodes 6.84\X faster than AR.

\subsection{Reasoning-Tuned Checkpoint}\label{app:thinking}

\begin{table*}[t]
\centering
\scriptsize
\setlength{\tabcolsep}{4pt}
\input{tables/tab_thinking}
\caption{Ouro-1.4B-Thinking on LoopSpec's thinking tasks. Speedup over own AR, mean of three repeats, and ratios of speedups, each the quotient of the two speedups it names. \methodH: the Ouro-1.4B draft head. Bold: largest speedup and largest ratio over LoopSpec per column and block. $D$: loops per output token relative to AR.}
\label{tab:thinking}
\end{table*}

Ouro-1.4B-Thinking runs LoopSpec's thinking tasks with LoopSpec's budgets of 4,096 to 16,384 tokens, and the AIME sets and MATH500 again with 32,768 tokens (\cref{tab:app-data}).
LoopSpec runs the faster of its proposal depths ``1'' and ``1:2'' per cell, and CopyLoop runs greedy with the configuration it uses for this checkpoint, $d_s=1$ and $K=2$ \citep{du2026copyloop}.
At LoopSpec's budgets, \method is faster than LoopSpec in all 8 cells and than CopyLoop in all 4 greedy cells (\cref{tab:thinking}).

At \Tone and LoopSpec's budgets, the ratio of \method over LoopSpec grows from \nThinkStart\X on GSM8K to \nThinkEnd\X on AIME25, as LoopSpec's speedup over its own AR falls on long outputs while that of \method holds at 2.96--3.20\X.
LoopSpec's schedule in our engine declines in the same way, from 2.13\X on GSM8K to 1.73\X on AIME25 over our AR.
On the H200, LoopSpec's official code reaches the mean accepted length that LoopSpec reports within 1\% in every cell at \Tone \citep[Table~2]{cho2026loopspec}.

\subsection{Engine Control}\label{app:engine}

Our AR is faster than SGLang's AR in four of the seven settings (\cref{tab:engine}, top).
To separate the method from the engine, we reimplemented LoopSpec's schedule in our engine from its released code, with its proposal depths, drafting and acceptance rule, captured CUDA graphs and at most one host synchronization per step.
We time it against the official code on 100 evenly spaced prompts of each main-table cell, with seed 0 at \Tone (\cref{tab:engine}, second block).
At the main table's configuration, the reimplementation takes LoopSpec's loops per output token within 3\% in every cell and runs at 0.94--1.08\X the speed of the official code in six settings, and at 1.44\X on Raven-OLMo, where our AR is also faster than SGLang's.
In the comparison below, where LoopSpec's schedule runs the configuration fastest in our engine per cell, \method is \nEngineFreeMin--\nEngineFreeMax\X as fast per model and mode. \methodH is \nEngineMin--\nEngineMax\X as fast, and faster in every cell and repeat.

\paragraph{Arms in our engine.}
On the same prompts, each cell times its arms in turn on one GPU, in three repeats (\cref{tab:engine}, third and fourth blocks).
In the arm without intermediate checks, the n-gram chain is checked only at admission and at full depth, on Raven under greedy decoding at its faster stride of 4 and, in \cref{tab:app-ablation}, also at stride 2.
In this arm, \method keeps \nLadXNMin--\nLadXNMax{} of its throughput per model and mode.

\paragraph{Without the pipeline.}
Run as ordinary speculative decoding, each pass runs every loop over the newest token and the draft tokens, which the full-depth readouts then check. The n-gram chain alone, at its fastest draft length per cell, then works as prompt lookup decoding \citep{saxena2023prompt} and reaches 1.64--2.27\X our AR. Both chains reach 1.90--2.73\X.
\subsection{Replacements per Check Depth}\label{app:depth}

On all prompts of \cref{app:engine}, we count the tokens each check depth of \method checks and replaces (\cref{tab:app-depth}).
At the admission check, a replacement is a draft token that the readout contradicts. At a later depth, it is the first token of a group that the readout contradicts, which may be a draft token, a token read out at the admission check or an earlier replacement.
Full depth checks every decoded token once.
The mean depth of the later replacements, \nMeanDepRvl{} of 32 loops on Raven-Llama under greedy decoding, places them close to the admission check. With the draft head, the pooled shares are nearly the same. Under greedy decoding, full depth keeps at least 94.3\% of the tokens that the first intermediate check of \method writes on Ouro and 99.0\% on Raven.

\subsection{Loops per Output Token}\label{app:loops}

\Cref{tab:app-loops} gives the two factors of \cref{eq:speedup} and the loops each decoder runs over all token positions. $D = S/(R\,N)$, with $S$ the loops run in sequence and $N$ the decoded tokens. $W$ counts the loops over all token positions of every pass, so that $W/(R\,N)$ is 1 for AR. Pooled over tasks, $1/(D \times \text{time per loop})$ differs from the geometric mean speedup of \cref{tab:main} by a factor of at most 1.17.
\methodH takes the fewest sequential loops per output token in all seven settings. \method takes fewer than LoopSpec in every setting.
On Raven-OLMo, LoopSpec's fastest configuration on MATH500 (4:8) takes 0.131 loops per output token. At LoopAhead's proposal depths, it takes 0.076 over all tasks in our calibration runs, close to the 0.078 that LoopAhead reports.
Per output token, \method and \methodH run 4.74--7.74\X as many loops as AR over all token positions. Their time per loop relative to their own AR is still 0.86--1.15 of LoopSpec's for \method and 0.96--1.27 for \methodH.

\subsection{Ablations}\label{app:ablations}

\Cref{tab:app-ablation} gives the \methodH entries of \cref{fig:mechanism}a.
The arm without intermediate checks runs \method on the prompts of \cref{app:engine} in all 33 cells (\cref{tab:engine}, third block).
Under greedy decoding it keeps \nLadXNOuro{} of the throughput on Ouro-1.4B and \nLadXNRaven{} on Raven-Llama, where it runs at its faster stride of 4 against the default of 2 with the checks. With both arms at stride 2, it keeps \nLadXtwoN{} on Raven-Llama.
At \Tone, with stride 4 in both arms on Raven, it keeps \nLadXNToneOuro{} on Ouro-1.4B and \nLadXNToneRaven{} on Raven-Llama (\cref{tab:app-ablation}).
A third arm draws the noise below full depth independently of full depth's, so a readout below full depth no longer shares the draws of the full-depth readout at its position. On Raven-Llama and the prompt subsets of \cref{app:engine}, it commits 0.45 tokens per step against 1.38 with the shared noise.

\subsection{Fewer Loops on Raven}\label{app:rscan}

Raven can run any number of loops at test time. We run Raven-Llama with $R=8$ and 16 loops as well as 32, greedy and at \Tone, on the prompts of \cref{app:engine} (\cref{tab:app-rscan}).
\methodH keeps its settings with the check depths above $R$ dropped, and LoopSpec's schedule in our engine runs its fastest configuration per cell and $R$.
The speedups of both decoders over our AR grow with $R$. Under greedy decoding, the ratio of \methodH over LoopSpec's schedule is 1.86--1.92\X.

\subsection{Huginn-0125}\label{app:huginn}

\begin{table}[t]
\centering
\scriptsize
\setlength{\tabcolsep}{3.2pt}
\input{tables/tab_huginn}
\caption{Huginn-0125 under greedy decoding. Speedup ($\times$) over AR in each method's engine, mean of three repeats. Geometric mean over tasks per repeat, then the mean. Lower block: without the prompts on which any method's output falls into a repetition loop. Bold: fastest per column and block.}
\label{tab:app-huginn}
\end{table}

Huginn-0125 \citep{geiping2025huginn} is a 3.5B recurrent-depth model trained from scratch with the same prelude, recurrent block and coda as the Raven models, which we run with $R=32$ loops. Its readouts after a few loops agree with full depth less often than those of the retrofitted Raven models, so we chose the stride of each decoder among 2, 4 and 8 on 782 sequences held out from the head's training data.
\method runs with stride $d=4$ and checks at depths 4, 8, 16 and 32, and \methodH with $d=8$ and checks at depths 8, 16 and 32. We trained the head on depth-8 states as in \cref{app:method}, for BBH without BBH questions.
LoopSpec's released server cannot load Huginn, so LoopSpec's schedule runs on CopyLoop's Huginn backend, which keeps LoopSpec's scheduling code.
Both baselines run their fastest configuration per task from a calibration as in \cref{app:baselines}. LoopSpec proposes at depths 4 and 8 on every task, and CopyLoop checks at depth 8 and copies 6 or 8 tokens.
Every method decodes all prompts of the five main tasks greedily on one H200, in three repeats (\cref{tab:app-huginn}). \method drafts up to $G=8$ tokens, as on the other models, and its speedup is over our AR, that of the baselines over SGLang's AR on CopyLoop's backend.
\method decodes \nHugFreeAR\X faster than AR, and \methodH \nHugHeadAR\X.
Over CopyLoop, the ratio of speedups of \method is \nHugFreeCL\X, and over LoopSpec it is \nHugFreeLS\X. Both decoders are faster than both baselines on every task and in every repeat, also without the prompts on which any method's output falls into a repetition loop (\cref{tab:app-huginn}, lower block).

\subsection{Batches and Long Contexts}\label{app:batch}

In a batched version of our engine, $B$ requests share each step's forward pass and a finished request is replaced by the next one. LoopSpec's released code decodes one request at a time, so we batched its schedule in this engine too.
Each cell decodes all prompts of \cref{app:engine} from GSM8K, MATH500 and MBPP+ greedily to the length of \methodH's output on one H200 and reports the decode throughput while all $B$ requests run.

A step holds 45 tokens per request on Ouro-1.4B and 153 on Raven-Llama, so a pass over many requests becomes compute-bound. We therefore set $G$, the draft chains and Raven's stride per batch size on the configuration prompts, and LoopSpec's configuration in the same way.
On Ouro-1.4B, $G=4$ at $B=4$ and 8, and $G=2$ with the head's chain alone at 16 and 32. On Raven-Llama, $G=4$ up to $B=8$, $G=2$ at 16 and 32, and stride $d=4$ from $B=8$.
With these settings, \methodH is at least 1.87\X faster than AR on every task and batch size (\cref{fig:mechanism}b).
Over the three tasks and 2 to 32 requests, it is 1.36--1.75\X faster than LoopSpec's batched schedule on Ouro-1.4B and 1.12--1.66\X on Raven-Llama.

\begin{table}[t]
\centering
\scriptsize
\setlength{\tabcolsep}{3.2pt}
\input{tables/tab_context}
\caption{Ratio of speedups of \methodH over each baseline on PG-19 contexts, greedy decoding with the end token ignored, mean of three repeats. Each speedup is over AR in the method's own engine, and each baseline runs its fastest configuration per length. Geo.\ mean: over lengths per repeat, then the mean.}
\label{tab:app-context}
\end{table}

\paragraph{Long contexts.}
On the PG-19 contexts of \cref{tab:app-data}, \methodH is faster than both baselines at every context length and in every repeat (\cref{tab:app-context}).
Over CopyLoop, the geometric mean of its ratio of speedups over lengths is 1.45\X to 1.66\X per model.
As every token of a step attends over the whole context, its ratio of speedups over LoopSpec falls to 1.30 at 32K on Ouro-1.4B.
Measured with one request at three of these lengths from 1K to 32K, its peak GPU memory, weights included, is 1.01--1.09\X AR's.
With 32 requests and the settings per batch size, it is 1.02--1.20\X AR's. These batches use contexts of 1K or 4K tokens.

\subsection{Accuracy and Output Identity}\label{app:accuracy}

\paragraph{FP32.}
\methodH and our AR run every operation in FP32, attention included, with TF32 off and the code otherwise unchanged.
On the prompts of \cref{app:engine} in all seven settings, these runs decode 845,772 tokens. In these FP32 runs, \methodH replaces a token at an intermediate readout 58,965 times and at full depth 4,699 times, and its outputs equal our AR's on all \nFPPrompts{} prompts.

\paragraph{BF16.}
\Cref{tab:app-acc} gives the accuracy of \methodH and our AR in every cell, measured as in \cref{app:protocol}.
Pooled over the greedy cells, 179 questions are right only for \methodH and 181 only for our AR. The pooled difference is \nGreedyDiff{} points, with a 95\% interval from \nGreedyDiffLo{} to \nGreedyDiffHi{} (sign-flip $p = \nGreedySignP$).
Pooled at \Tone, it is \nToneDiff{} points, from \nToneDiffLo{} to \nToneDiffHi{} (sign-flip $p = \nToneSignP$).
Against the FP32 output, which \methodH and our AR share on every prompt of \cref{app:engine}, the BF16 output of \methodH differs on 39.5\% of these 3,300 prompts. That of our AR differs on 39.0\% (exact McNemar $p = \nBFRefP$).
Between decoders, we compare each pair of outputs up to the shorter of the two, on all 5,802 greedy prompts of Ouro's main tasks (first repeat).
The output of \methodH equals our AR's on \nBFOurs\% of them and that of \method on 74.5\%. LoopSpec and CopyLoop equal the SGLang AR of their own server on \nBFLS\% and \nBFCL\%.
Our AR equals SGLang's AR on \nBFAR\%, while the SGLang ARs of the two servers agree on every prompt, so these differences come from the kernels and the shapes of the passes.
Ouro-2.6B scores below Ouro-1.4B on MATH500 with every decoder, SGLang's AR in the code of LoopSpec and CopyLoop included (56.0\% and 56.4\%), as it follows the short exemplars of the prompt.

\subsection{RTX 5090}\label{app:gpu}

On RTX 5090 GPUs, we chose every method's settings again (\cref{tab:app-gpu}). Rescoring our settings on the configuration prompts changed only the head's draws at \Tone on Raven, to $0.7\times T$. Each cell ran all methods, each baseline at its fastest candidate, on the prompts of \cref{app:engine}, in three repeats.
The ratio of speedups of \methodH over LoopSpec is \nGPUMin--\nGPUMax\X per model and mode.
Under greedy decoding, the ratio of speedups of \method over CopyLoop is \nGPUFreeCLMin--\nGPUFreeCLMax\X. That of \methodH is 1.11--1.60\X.

%% file: tables/tab_heldout.tex
\begin{tabular}{@{}ll ccccc@{\hspace{6pt}}c ccccc@{\hspace{6pt}}c@{}}
\toprule
 & & \multicolumn{6}{c}{Over LoopSpec} & \multicolumn{6}{c}{Over CopyLoop} \\
\cmidrule(lr){3-8}\cmidrule(l){9-14}
 & & \multicolumn{5}{c}{Held-out datasets} & & \multicolumn{5}{c}{Held-out datasets} & \\
\cmidrule(lr){3-7}\cmidrule(lr){9-13}
Model & Method & SVAMP & Olymp. & MMLU-Pro & BCB & Mean & Spec-Bench & SVAMP & Olymp. & MMLU-Pro & BCB & Mean & Spec-Bench \\
\midrule
\multicolumn{14}{c}{\textit{Greedy}} \\
\midrule
\multirow{2}{*}{Ouro-1.4B} & \method & 1.36 & 1.26 & 1.39 & 2.27 & \cellcolor{ours}1.52 & 1.46 & 1.10 & 1.15 & 1.42 & 1.35 & \cellcolor{ours}1.25 & 1.18 \\
 & \methodH & 1.71 & 1.68 & 1.44 & 2.41 & \cellcolor{ours}1.78 & 1.49 & 1.38 & 1.53 & 1.47 & 1.43 & \cellcolor{ours}1.45 & 1.20 \\
\multirow{2}{*}{Ouro-2.6B} & \method & 1.36 & 1.84 & 1.26 & 2.13 & \cellcolor{ours}1.61 & 1.46 & 0.99 & 1.23 & 1.12 & 1.21 & \cellcolor{ours}1.13 & 1.22 \\
 & \methodH & 1.86 & 2.25 & 1.42 & 2.34 & \cellcolor{ours}1.93 & 1.54 & 1.36 & 1.51 & 1.25 & 1.33 & \cellcolor{ours}1.36 & 1.28 \\
\multirow{2}{*}{Raven-Llama} & \method & 1.29 & 2.34 & 2.01 & 2.53 & \cellcolor{ours}1.98 & 2.66 & 1.10 & 1.31 & 1.43 & 1.39 & \cellcolor{ours}1.30 & 1.49 \\
 & \methodH & 1.43 & 2.57 & 2.08 & 2.62 & \cellcolor{ours}2.12 & 2.72 & 1.22 & 1.44 & 1.48 & 1.44 & \cellcolor{ours}1.39 & 1.53 \\
\multirow{2}{*}{Raven-OLMo} & \method & 1.43 & 2.23 & 2.47 & 2.77 & \cellcolor{ours}2.16 & 2.45 & 1.19 & 1.57 & 1.74 & 1.54 & \cellcolor{ours}1.50 & 1.45 \\
 & \methodH & 1.75 & 2.46 & 2.44 & 2.77 & \cellcolor{ours}2.32 & 2.52 & 1.46 & 1.74 & 1.73 & 1.54 & \cellcolor{ours}1.61 & 1.49 \\
\multirow{2}{*}{Raven-TinyLlama} & \method & 1.45 & -- & -- & 2.38 & \cellcolor{ours}1.86 & 2.13 & 1.25 & -- & -- & 1.41 & \cellcolor{ours}1.33 & 1.35 \\
 & \methodH & 1.71 & -- & -- & 2.51 & \cellcolor{ours}2.07 & 2.28 & 1.47 & -- & -- & 1.48 & \cellcolor{ours}1.47 & 1.44 \\
\midrule
\multicolumn{14}{c}{\textit{\Tone, top-p 0.9}} \\
\midrule
\multirow{2}{*}{Ouro-1.4B} & \method & 1.36 & 1.27 & 1.16 & 2.13 & \cellcolor{ours}1.44 & 1.22 & -- & -- & -- & -- & \cellcolor{ours}-- & -- \\
 & \methodH & 1.60 & 1.55 & 1.24 & 2.21 & \cellcolor{ours}1.61 & 1.23 & -- & -- & -- & -- & \cellcolor{ours}-- & -- \\
\multirow{2}{*}{Raven-Llama} & \method & 1.35 & 1.44 & 1.37 & 1.76 & \cellcolor{ours}1.47 & 1.27 & -- & -- & -- & -- & \cellcolor{ours}-- & -- \\
 & \methodH & 1.53 & 1.76 & 1.45 & 1.93 & \cellcolor{ours}1.66 & 1.37 & -- & -- & -- & -- & \cellcolor{ours}-- & -- \\
\bottomrule
\end{tabular}

%% file: tables/tab_main_norep.tex
\begin{tabular}{@{}ll cccccc@{}}
\toprule
Model & Method & GSM8K & MATH500 & BBH & HumanEval+ & MBPP+ & Geo.\ mean \\
\midrule
\multicolumn{8}{c}{\textit{Greedy decoding}} \\
\midrule
 & LoopSpec & 2.88{\tiny$\pm$0.07} & 3.12{\tiny$\pm$0.06} & 2.78{\tiny$\pm$0.06} & 3.19{\tiny$\pm$0.15} & 2.86{\tiny$\pm$0.01} & 2.96{\tiny$\pm$0.03} \\
 & CopyLoop & 3.38{\tiny$\pm$0.20} {\scriptsize(1.17)} & 3.22{\tiny$\pm$0.11} {\scriptsize(1.03)} & 3.20{\tiny$\pm$0.12} {\scriptsize(1.15)} & 5.49{\tiny$\pm$0.18} {\scriptsize(1.72)} & 2.68{\tiny$\pm$0.10} {\scriptsize(0.94)} & 3.48{\tiny$\pm$0.08} {\scriptsize(1.18)} \\
 & \cellcolor{ours}\method & \cellcolor{ours}3.80{\tiny$\pm$0.02} {\scriptsize(1.32)} & \cellcolor{ours}3.73{\tiny$\pm$0.14} {\scriptsize(1.20)} & \cellcolor{ours}\textbf{5.43}{\tiny$\pm$0.17} {\scriptsize(1.95)} & \cellcolor{ours}6.90{\tiny$\pm$0.22} {\scriptsize(2.16)} & \cellcolor{ours}3.10{\tiny$\pm$0.02} {\scriptsize(1.08)} & \multicolumn{1}{>{\columncolor{ours}[\tabcolsep][0pt]}c@{}}{4.40{\tiny$\pm$0.06} {\scriptsize(1.49)}} \\
\multirow{-4}{*}{\makecell[l]{Ouro-\\1.4B}} & \cellcolor{ours}\enspace + head & \cellcolor{ours}\textbf{5.01}{\tiny$\pm$0.02} {\scriptsize(1.74)} & \cellcolor{ours}\textbf{5.16}{\tiny$\pm$0.20} {\scriptsize(1.65)} & \cellcolor{ours}5.19{\tiny$\pm$0.13} {\scriptsize(1.87)} & \cellcolor{ours}\textbf{8.57}{\tiny$\pm$0.16} {\scriptsize(2.69)} & \cellcolor{ours}\textbf{4.31}{\tiny$\pm$0.12} {\scriptsize(1.51)} & \multicolumn{1}{>{\columncolor{ours}[\tabcolsep][0pt]}c@{}}{\textbf{5.48}{\tiny$\pm$0.08} {\scriptsize(1.85)}} \\
\midrule
 & LoopSpec & 3.27{\tiny$\pm$0.04} & 3.27{\tiny$\pm$0.05} & 2.94{\tiny$\pm$0.11} & 3.23{\tiny$\pm$0.04} & 3.17{\tiny$\pm$0.14} & 3.17{\tiny$\pm$0.04} \\
 & CopyLoop & 4.27{\tiny$\pm$0.02} {\scriptsize(1.31)} & 3.97{\tiny$\pm$0.02} {\scriptsize(1.21)} & 4.13{\tiny$\pm$0.09} {\scriptsize(1.40)} & 6.26{\tiny$\pm$0.02} {\scriptsize(1.94)} & 3.33{\tiny$\pm$0.06} {\scriptsize(1.05)} & 4.29{\tiny$\pm$0.02} {\scriptsize(1.35)} \\
 & \cellcolor{ours}\method & \cellcolor{ours}4.34{\tiny$\pm$0.26} {\scriptsize(1.33)} & \cellcolor{ours}4.47{\tiny$\pm$0.18} {\scriptsize(1.37)} & \cellcolor{ours}\textbf{6.35}{\tiny$\pm$0.15} {\scriptsize(2.16)} & \cellcolor{ours}7.10{\tiny$\pm$0.02} {\scriptsize(2.20)} & \cellcolor{ours}3.23{\tiny$\pm$0.07} {\scriptsize(1.02)} & \multicolumn{1}{>{\columncolor{ours}[\tabcolsep][0pt]}c@{}}{4.90{\tiny$\pm$0.10} {\scriptsize(1.55)}} \\
\multirow{-4}{*}{\makecell[l]{Ouro-\\2.6B}} & \cellcolor{ours}\enspace + head & \cellcolor{ours}\textbf{6.12}{\tiny$\pm$0.02} {\scriptsize(1.87)} & \cellcolor{ours}\textbf{5.85}{\tiny$\pm$0.03} {\scriptsize(1.79)} & \cellcolor{ours}6.04{\tiny$\pm$0.15} {\scriptsize(2.05)} & \cellcolor{ours}\textbf{8.77}{\tiny$\pm$0.05} {\scriptsize(2.72)} & \cellcolor{ours}\textbf{4.72}{\tiny$\pm$0.07} {\scriptsize(1.49)} & \multicolumn{1}{>{\columncolor{ours}[\tabcolsep][0pt]}c@{}}{\textbf{6.17}{\tiny$\pm$0.05} {\scriptsize(1.95)}} \\
\midrule
 & LoopSpec & 4.17{\tiny$\pm$0.07} & 5.57{\tiny$\pm$0.13} & 5.52{\tiny$\pm$0.25} & 6.01{\tiny$\pm$0.14} & 4.99{\tiny$\pm$0.10} & 5.21{\tiny$\pm$0.05} \\
 & CopyLoop & 4.99{\tiny$\pm$0.16} {\scriptsize(1.20)} & 6.42{\tiny$\pm$0.13} {\scriptsize(1.15)} & 8.40{\tiny$\pm$0.19} {\scriptsize(1.52)} & 9.38{\tiny$\pm$0.41} {\scriptsize(1.56)} & 5.51{\tiny$\pm$0.07} {\scriptsize(1.10)} & 6.74{\tiny$\pm$0.10} {\scriptsize(1.29)} \\
 & \cellcolor{ours}\method & \cellcolor{ours}5.68{\tiny$\pm$0.34} {\scriptsize(1.36)} & \cellcolor{ours}7.98{\tiny$\pm$0.22} {\scriptsize(1.43)} & \cellcolor{ours}\textbf{11.08}{\tiny$\pm$0.29} {\scriptsize(2.01)} & \cellcolor{ours}12.53{\tiny$\pm$0.38} {\scriptsize(2.08)} & \cellcolor{ours}7.21{\tiny$\pm$0.24} {\scriptsize(1.44)} & \multicolumn{1}{>{\columncolor{ours}[\tabcolsep][0pt]}c@{}}{8.54{\tiny$\pm$0.07} {\scriptsize(1.64)}} \\
\multirow{-4}{*}{\makecell[l]{Raven-\\Llama}} & \cellcolor{ours}\enspace + head & \cellcolor{ours}\textbf{6.23}{\tiny$\pm$0.09} {\scriptsize(1.49)} & \cellcolor{ours}\textbf{9.28}{\tiny$\pm$0.37} {\scriptsize(1.67)} & \cellcolor{ours}10.24{\tiny$\pm$0.29} {\scriptsize(1.86)} & \cellcolor{ours}\textbf{13.80}{\tiny$\pm$0.42} {\scriptsize(2.30)} & \cellcolor{ours}\textbf{8.05}{\tiny$\pm$0.10} {\scriptsize(1.61)} & \multicolumn{1}{>{\columncolor{ours}[\tabcolsep][0pt]}c@{}}{\textbf{9.20}{\tiny$\pm$0.11} {\scriptsize(1.77)}} \\
\midrule
 & LoopSpec & 5.17{\tiny$\pm$0.02} & 4.40{\tiny$\pm$0.07} & 4.76{\tiny$\pm$0.06} & 6.03{\tiny$\pm$0.08} & 5.97{\tiny$\pm$0.02} & 5.23{\tiny$\pm$0.02} \\
 & CopyLoop & 6.37{\tiny$\pm$0.23} {\scriptsize(1.23)} & 5.14{\tiny$\pm$0.06} {\scriptsize(1.17)} & 6.81{\tiny$\pm$0.14} {\scriptsize(1.43)} & 9.24{\tiny$\pm$0.13} {\scriptsize(1.53)} & 9.19{\tiny$\pm$0.37} {\scriptsize(1.54)} & 7.17{\tiny$\pm$0.14} {\scriptsize(1.37)} \\
 & \cellcolor{ours}\method & \cellcolor{ours}8.17{\tiny$\pm$0.45} {\scriptsize(1.58)} & \cellcolor{ours}6.78{\tiny$\pm$0.21} {\scriptsize(1.54)} & \cellcolor{ours}\textbf{9.62}{\tiny$\pm$0.55} {\scriptsize(2.02)} & \cellcolor{ours}12.71{\tiny$\pm$0.21} {\scriptsize(2.11)} & \cellcolor{ours}11.26{\tiny$\pm$0.42} {\scriptsize(1.89)} & \multicolumn{1}{>{\columncolor{ours}[\tabcolsep][0pt]}c@{}}{9.47{\tiny$\pm$0.28} {\scriptsize(1.81)}} \\
\multirow{-4}{*}{\makecell[l]{Raven-\\OLMo}} & \cellcolor{ours}\enspace + head & \cellcolor{ours}\textbf{9.73}{\tiny$\pm$0.04} {\scriptsize(1.88)} & \cellcolor{ours}\textbf{8.18}{\tiny$\pm$0.04} {\scriptsize(1.86)} & \cellcolor{ours}9.22{\tiny$\pm$0.31} {\scriptsize(1.94)} & \cellcolor{ours}\textbf{14.22}{\tiny$\pm$0.26} {\scriptsize(2.36)} & \cellcolor{ours}\textbf{13.35}{\tiny$\pm$0.44} {\scriptsize(2.24)} & \multicolumn{1}{>{\columncolor{ours}[\tabcolsep][0pt]}c@{}}{\textbf{10.69}{\tiny$\pm$0.04} {\scriptsize(2.04)}} \\
\midrule
 & LoopSpec & 5.15{\tiny$\pm$0.20} & -- & -- & 6.46{\tiny$\pm$0.14} & 5.61{\tiny$\pm$0.17} & 5.71{\tiny$\pm$0.06} \\
 & CopyLoop & 5.68{\tiny$\pm$0.14} {\scriptsize(1.10)} & -- & -- & 9.78{\tiny$\pm$0.30} {\scriptsize(1.51)} & 6.36{\tiny$\pm$0.31} {\scriptsize(1.13)} & 7.07{\tiny$\pm$0.01} {\scriptsize(1.24)} \\
 & \cellcolor{ours}\method & \cellcolor{ours}7.86{\tiny$\pm$0.41} {\scriptsize(1.53)} & \cellcolor{ours}-- & \cellcolor{ours}-- & \cellcolor{ours}14.62{\tiny$\pm$0.30} {\scriptsize(2.26)} & \cellcolor{ours}8.77{\tiny$\pm$0.14} {\scriptsize(1.56)} & \multicolumn{1}{>{\columncolor{ours}[\tabcolsep][0pt]}c@{}}{10.03{\tiny$\pm$0.19} {\scriptsize(1.76)}} \\
\multirow{-4}{*}{\makecell[l]{Raven-\\TinyLlama}} & \cellcolor{ours}\enspace + head & \cellcolor{ours}\textbf{9.01}{\tiny$\pm$0.17} {\scriptsize(1.75)} & \cellcolor{ours}-- & \cellcolor{ours}-- & \cellcolor{ours}\textbf{15.76}{\tiny$\pm$0.16} {\scriptsize(2.44)} & \cellcolor{ours}\textbf{10.30}{\tiny$\pm$0.12} {\scriptsize(1.84)} & \multicolumn{1}{>{\columncolor{ours}[\tabcolsep][0pt]}c@{}}{\textbf{11.35}{\tiny$\pm$0.11} {\scriptsize(1.99)}} \\
\midrule
\multicolumn{8}{c}{\textit{Sampling, \Tone, top-p 0.9}} \\
\midrule
 & LoopSpec & 2.34{\tiny$\pm$0.01} & 2.39{\tiny$\pm$0.01} & 2.25{\tiny$\pm$0.01} & 2.59{\tiny$\pm$0.06} & 2.43{\tiny$\pm$0.08} & 2.40{\tiny$\pm$0.02} \\
 & \cellcolor{ours}\method & \cellcolor{ours}3.07{\tiny$\pm$0.04} {\scriptsize(1.31)} & \cellcolor{ours}3.48{\tiny$\pm$0.18} {\scriptsize(1.46)} & \cellcolor{ours}\textbf{4.54}{\tiny$\pm$0.08} {\scriptsize(2.02)} & \cellcolor{ours}5.16{\tiny$\pm$0.07} {\scriptsize(1.99)} & \cellcolor{ours}3.03{\tiny$\pm$0.10} {\scriptsize(1.25)} & \multicolumn{1}{>{\columncolor{ours}[\tabcolsep][0pt]}c@{}}{3.77{\tiny$\pm$0.09} {\scriptsize(1.57)}} \\
\multirow{-3}{*}{\makecell[l]{Ouro-\\1.4B}} & \cellcolor{ours}\enspace + head & \cellcolor{ours}\textbf{3.66}{\tiny$\pm$0.01} {\scriptsize(1.56)} & \cellcolor{ours}\textbf{4.35}{\tiny$\pm$0.00} {\scriptsize(1.82)} & \cellcolor{ours}4.31{\tiny$\pm$0.01} {\scriptsize(1.92)} & \cellcolor{ours}\textbf{5.77}{\tiny$\pm$0.06} {\scriptsize(2.23)} & \cellcolor{ours}\textbf{3.74}{\tiny$\pm$0.07} {\scriptsize(1.54)} & \multicolumn{1}{>{\columncolor{ours}[\tabcolsep][0pt]}c@{}}{\textbf{4.31}{\tiny$\pm$0.01} {\scriptsize(1.80)}} \\
\midrule
 & LoopSpec & 3.54{\tiny$\pm$0.06} & 4.12{\tiny$\pm$0.02} & 4.10{\tiny$\pm$0.01} & 3.99{\tiny$\pm$0.07} & 3.78{\tiny$\pm$0.03} & 3.90{\tiny$\pm$0.01} \\
 & \cellcolor{ours}\method & \cellcolor{ours}4.90{\tiny$\pm$0.14} {\scriptsize(1.38)} & \cellcolor{ours}6.14{\tiny$\pm$0.27} {\scriptsize(1.49)} & \cellcolor{ours}\textbf{8.38}{\tiny$\pm$0.40} {\scriptsize(2.04)} & \cellcolor{ours}6.86{\tiny$\pm$0.01} {\scriptsize(1.72)} & \cellcolor{ours}5.79{\tiny$\pm$0.03} {\scriptsize(1.53)} & \multicolumn{1}{>{\columncolor{ours}[\tabcolsep][0pt]}c@{}}{6.31{\tiny$\pm$0.16} {\scriptsize(1.62)}} \\
\multirow{-3}{*}{\makecell[l]{Raven-\\Llama}} & \cellcolor{ours}\enspace + head & \cellcolor{ours}\textbf{5.64}{\tiny$\pm$0.02} {\scriptsize(1.59)} & \cellcolor{ours}\textbf{7.17}{\tiny$\pm$0.05} {\scriptsize(1.74)} & \cellcolor{ours}8.04{\tiny$\pm$0.08} {\scriptsize(1.96)} & \cellcolor{ours}\textbf{7.79}{\tiny$\pm$0.01} {\scriptsize(1.95)} & \cellcolor{ours}\textbf{6.75}{\tiny$\pm$0.02} {\scriptsize(1.79)} & \multicolumn{1}{>{\columncolor{ours}[\tabcolsep][0pt]}c@{}}{\textbf{7.02}{\tiny$\pm$0.02} {\scriptsize(1.80)}} \\
\bottomrule
\end{tabular}

%% file: tables/tab_engine.tex
\begin{tabular}{@{}l ccccc cc@{}}
\toprule
 & \multicolumn{5}{c}{Greedy} & \multicolumn{2}{c}{\Tone, top-p 0.9} \\
\cmidrule(lr){2-6}\cmidrule(l){7-8}
 & \makecell{Ouro-\\1.4B} & \makecell{Ouro-\\2.6B} & \makecell{Raven-\\Llama} & \makecell{Raven-\\OLMo} & \makecell{Raven-\\TinyLlama} & \makecell{Ouro-\\1.4B} & \makecell{Raven-\\Llama} \\
\midrule
\multicolumn{8}{c}{\textit{Raw decode throughput (tok/s) in \cref{tab:main}}} \\
\midrule
SGLang's AR & 141.9 & 70.8 & 68.5 & 42.3 & 62.8 & 144.1 & 68.8 \\
Our AR & 164.2 & 83.3 & 67.1 & 60.3 & 58.9 & 162.0 & 66.1 \\
LoopSpec & 420.5 & 226.5 & 380.3 & 222.4 & 373.5 & 345.1 & 268.6 \\
\method & 728.8 & 433.6 & 692.6 & 700.0 & 648.2 & 608.5 & 416.5 \\
\methodH & 906.3 & 545.0 & 744.1 & 770.3 & 730.4 & 695.7 & 462.9 \\
\midrule
\multicolumn{8}{c}{\textit{LoopSpec's schedule in our engine against its official code, 100 prompts per main-table cell}} \\
\midrule
LoopSpec, official code (tok/s) & 418.3 & 233.2 & 379.4 & 226.7 & 383.1 & 348.3 & 268.3 \\
LoopSpec, our engine (tok/s) & 452.8 & 248.1 & 392.8 & 325.6 & 404.2 & 354.4 & 253.0 \\
LoopSpec, our engine / official code & 1.08 & 1.06 & 1.04 & 1.44 & 1.06 & 1.02 & 0.94 \\
\midrule
\multicolumn{8}{c}{\textit{Our engine, 100 prompts per main-table cell, over \method}} \\
\midrule
\method without intermediate checks$^{*}$ & 0.888 & 0.918 & 0.754 & 0.794 & 0.734 & 0.862 & 0.792 \\
\midrule
\multicolumn{8}{c}{\textit{Our engine, over LoopSpec's schedule}} \\
\midrule
\method & 1.64 & 1.78 & 1.73 & 2.08 & 1.66 & 1.73 & 1.67 \\
\methodH & 2.00 & 2.28 & 1.88 & 2.35 & 1.85 & 1.99 & 1.88 \\
\midrule
\multicolumn{8}{c}{\textit{Speculative decoding without the pipeline, same prompts, over our AR}} \\
\midrule
N-gram chain alone & 1.93 & -- & 2.27 & -- & -- & 1.69 & 1.64 \\
Both chains & 2.59 & -- & 2.73 & -- & -- & 2.20 & 1.90 \\
\bottomrule
\end{tabular}

%% file: tables/tab_depth.tex
\begin{tabular}{@{}l rrrrrr@{}}
\toprule
 & \makecell{Admis-\\sion\\check:\\rejected\\(\%)} & \makecell{Inter-\\mediate\\depths:\\replaced} & \makecell{Full\\depth:\\replaced} & \makecell{Below\\full\\depth\\(\%)} & \makecell{Full\\depth:\\rate\\(\%)} & \makecell{Mean\\depth} \\
\midrule
\multicolumn{7}{@{}l}{\textit{Greedy}} \\
Ouro-1.4B & 33.6 & 9,493 & 1,247 & 88.4 & 1.10 & 2.5 \\
Ouro-2.6B & 25.3 & 5,759 & 719 & 88.9 & 0.69 & 2.5 \\
Raven-Llama & 19.9 & 6,897 & 207 & 97.1 & 0.16 & 6.7 \\
Raven-OLMo & 18.8 & 13,582 & 387 & 97.2 & 0.18 & 6.4 \\
Raven-TinyLlama & 22.4 & 2,733 & 72 & 97.4 & 0.14 & 6.6 \\
All & 22.9 & 38,464 & 2,632 & 93.6 & 0.43 & -- \\
All + head & 28.1 & 37,757 & 2,583 & 93.6 & 0.42 & -- \\
\midrule
\multicolumn{7}{@{}l}{\textit{\Tone}} \\
Ouro-1.4B & 38.2 & 15,131 & 2,413 & 86.2 & 2.22 & 2.5 \\
Raven-Llama & 31.4 & 6,037 & 655 & 90.2 & 0.52 & 12.1 \\
All & 34.8 & 21,168 & 3,068 & 87.3 & 1.30 & -- \\
All + head & 40.7 & 21,007 & 2,832 & 88.1 & 1.20 & -- \\
\bottomrule
\end{tabular}

%% file: tables/tab_loops.tex
\begin{tabular}{@{}l ccccc cc@{}}
\toprule
 & \multicolumn{5}{c}{Greedy} & \multicolumn{2}{c}{\Tone, top-p 0.9} \\
\cmidrule(lr){2-6}\cmidrule(l){7-8}
 & \makecell{Ouro-\\1.4B} & \makecell{Ouro-\\2.6B} & \makecell{Raven-\\Llama} & \makecell{Raven-\\OLMo} & \makecell{Raven-\\TinyLlama} & \makecell{Ouro-\\1.4B} & \makecell{Raven-\\Llama} \\
\midrule
\multicolumn{8}{c}{\textit{Loops per output token, $D$}} \\
\midrule
LoopSpec & 0.2853 & 0.2725 & 0.0773 & 0.0954 & 0.0840 & 0.3032 & 0.1291 \\
CopyLoop & 0.1984 & 0.1475 & 0.0633 & 0.0619 & 0.0815 & -- & -- \\
\method & 0.1833 & 0.1498 & 0.0395 & 0.0380 & 0.0511 & 0.2151 & 0.0961 \\
\methodH & \textbf{0.1299} & \textbf{0.1078} & \textbf{0.0321} & \textbf{0.0310} & \textbf{0.0399} & \textbf{0.1628} & \textbf{0.0769} \\
\midrule
\multicolumn{8}{c}{\textit{Tokens committed per step, $\tau = d/(R\,D)$}} \\
\midrule
\method & 1.36 & 1.67 & 1.58 & 1.64 & 1.22 & 1.16 & 1.30 \\
\methodH & 1.92 & 2.32 & 1.95 & 2.02 & 1.57 & 1.54 & 1.63 \\
\midrule
\multicolumn{8}{c}{\textit{Time per loop relative to own AR}} \\
\midrule
LoopSpec & \textbf{1.17} & \textbf{1.13} & \textbf{2.27} & \textbf{2.16} & 2.11 & 1.40 & 1.99 \\
\method & 1.31 & 1.25 & 2.37 & 2.48 & \textbf{2.09} & \textbf{1.31} & \textbf{1.72} \\
\methodH & 1.46 & 1.36 & 2.77 & 2.75 & 2.29 & 1.47 & 1.91 \\
\midrule
\multicolumn{8}{c}{\textit{Loops over all token positions per output token, $W/(R\,N)$}} \\
\midrule
LoopSpec in our engine & \textbf{1.11} & \textbf{1.07} & \textbf{1.32} & \textbf{1.22} & \textbf{1.31} & \textbf{1.27} & \textbf{1.99} \\
\method & 6.60 & 5.39 & 5.69 & 5.47 & 7.36 & 7.74 & 6.92 \\
\methodH & 5.85 & 4.85 & 4.91 & 4.74 & 6.10 & 7.33 & 6.23 \\
\midrule
\multicolumn{8}{c}{\textit{As LoopAhead reports (read off its figures), its LoopSpec in brackets}} \\
\midrule
LoopAhead $D$ & 0.16 (0.29) & 0.15 (0.27) & 0.085 (0.076) & 0.083 (0.078) & -- & -- & -- \\
LoopAhead $W/(R\,N)$ & 1.91 (1.14) & 1.82 (1.09) & 1.16 (1.41) & 1.14 (1.64) & -- & -- & -- \\
\bottomrule
\end{tabular}

%% file: tables/tab_app_ablation.tex
\begin{tabular}{@{}l cccc@{}}
\toprule
 & \multicolumn{2}{c}{Ouro-1.4B} & \multicolumn{2}{c}{Raven-Llama} \\
\cmidrule(lr){2-3}\cmidrule(l){4-5}
Removed component & Greedy & \Tone & Greedy & \Tone \\
\midrule
Multi-token admission ($G=0$) & 0.521 {\scriptsize(0.512--0.526)} & 0.624 {\scriptsize(0.618--0.633)} & 0.585 {\scriptsize(0.582--0.589)} & 0.681 {\scriptsize(0.667--0.693)} \\
Admission check$^{*}$ & 0.606 {\scriptsize(0.603--0.608)} & 0.629 {\scriptsize(0.621--0.638)} & 0.582 {\scriptsize(0.577--0.586)} & 0.559 {\scriptsize(0.543--0.583)} \\
N-gram chain & 0.731 {\scriptsize(0.726--0.741)} & 0.793 {\scriptsize(0.787--0.803)} & 0.681 {\scriptsize(0.675--0.692)} & 0.735 {\scriptsize(0.715--0.757)} \\
Redrafting after a replacement & 0.984 {\scriptsize(0.972--0.994)} & 0.973 {\scriptsize(0.968--0.981)} & 0.982 {\scriptsize(0.980--0.986)} & 0.981 {\scriptsize(0.975--0.991)} \\
\midrule
\multicolumn{5}{@{}l}{\textit{\method}} \\
Intermediate checks$^{\dagger}$ & 0.888 {\scriptsize(0.875--0.895)} & 0.862 {\scriptsize(0.856--0.870)} & 0.754 {\scriptsize(0.748--0.766)} & 0.792 {\scriptsize(0.787--0.800)} \\
Intermediate checks, stride 2$^{\ddagger}$ & -- & -- & 0.552 {\scriptsize(0.548--0.555)} & -- \\
Shared noise below full depth$^{\S}$ & -- & 0.764 {\scriptsize(0.757--0.768)} & -- & 0.331 {\scriptsize(0.326--0.341)} \\
\bottomrule
\end{tabular}

%% file: tables/tab_rscan.tex
\begin{tabular}{@{}lr ccc@{}}
\toprule
Mode & $R$ & \makecell{\methodH\\/ our AR} & \makecell{LoopSpec\\/ our AR} & \makecell{\methodH\\/ LoopSpec} \\
\midrule
Greedy & 8 & 4.52 & 2.36 & 1.92 \\
 & 16 & 6.97 & 3.67 & 1.90 \\
 & 32 & 10.91 & 5.88 & 1.86 \\
\midrule
\Tone & 8 & 2.52 & 1.59 & 1.58 \\
 & 16 & 4.06 & 2.43 & 1.67 \\
 & 32 & 6.97 & 3.75 & 1.86 \\
\bottomrule
\end{tabular}

%% file: tables/tab_acc.tex
\begin{tabular}{@{}l ccccc ccccc@{}}
\toprule
 & \multicolumn{5}{c}{Our AR} & \multicolumn{5}{c}{\methodH} \\
\cmidrule(lr){2-6}\cmidrule(l){7-11}
Model & GSM8K & MATH500 & BBH & HE+ & MBPP+ & GSM8K & MATH500 & BBH & HE+ & MBPP+ \\
\midrule
Ouro-1.4B & 78.5 & 72.6 & 69.6 & 65.9 & 64.8 & 79.1 & 73.2 & 70.7 & 66.5 & 64.8 \\
Ouro-2.6B & 86.0 & 57.4 & 77.8 & 73.2 & 67.7 & 86.2 & 56.0 & 77.4 & 72.6 & 69.0 \\
Raven-Llama & 51.9 & 35.4 & 32.8 & 11.0 & 23.8 & 52.0 & 35.8 & 31.5 & 9.8 & 24.6 \\
Raven-OLMo & 46.1 & 28.0 & 28.0 & 10.4 & 12.2 & 45.3 & 27.6 & 28.7 & 11.0 & 11.6 \\
Raven-TinyLlama & 43.7 & -- & -- & 14.0 & 23.0 & 43.1 & -- & -- & 14.0 & 23.8 \\
\midrule
Ouro-1.4B, \Tone & 67.0 & 62.9 & 62.8 & 57.3 & 53.6 & 67.4 & 62.3 & 62.5 & 57.7 & 53.5 \\
Raven-Llama, \Tone & 36.4 & 21.5 & 33.5 & 5.9 & 10.2 & 35.5 & 20.9 & 33.5 & 5.7 & 10.4 \\
\bottomrule
\end{tabular}

%% file: tables/tab_gpu.tex
\begin{tabular}{@{}l ccccc cc@{}}
\toprule
 & \multicolumn{5}{c}{Greedy} & \multicolumn{2}{c}{\Tone, top-p 0.9} \\
\cmidrule(lr){2-6}\cmidrule(l){7-8}
 & \makecell{Ouro-\\1.4B} & \makecell{Ouro-\\2.6B} & \makecell{Raven-\\Llama} & \makecell{Raven-\\OLMo} & \makecell{Raven-\\TinyLlama} & \makecell{Ouro-\\1.4B} & \makecell{Raven-\\Llama} \\
\midrule
LoopSpec & 2.59 & 2.75 & 5.21 & 5.03 & 6.07 & 2.26 & 4.05 \\
CopyLoop & 4.00 {\scriptsize(1.54)} & 4.81 {\scriptsize(1.75)} & 8.28 {\scriptsize(1.59)} & 8.75 {\scriptsize(1.74)} & 8.76 {\scriptsize(1.44)} & -- & -- \\
\method & 4.78 {\scriptsize(1.85)} & 5.64 {\scriptsize(2.05)} & 7.96 {\scriptsize(1.53)} & 10.69 {\scriptsize(2.13)} & 9.35 {\scriptsize(1.54)} & 3.94 {\scriptsize(1.74)} & 5.49 {\scriptsize(1.36)} \\
\enspace + head & \textbf{6.39} {\scriptsize(2.47)} & \textbf{7.65} {\scriptsize(2.78)} & \textbf{9.17} {\scriptsize(1.76)} & \textbf{13.17} {\scriptsize(2.62)} & \textbf{11.17} {\scriptsize(1.84)} & \textbf{4.91} {\scriptsize(2.17)} & \textbf{6.34} {\scriptsize(1.57)} \\
\midrule
LoopSpec in our engine & 2.77 & 2.91 & 5.22 & 5.47 & 6.19 & 2.35 & 3.84 \\
\bottomrule
\end{tabular}

%% file: tables/tab_thinking.tex
\begin{tabular}{@{}l cccc ccc@{}}
\toprule
 & \multicolumn{4}{c}{Budgets of 4,096--16,384 tokens} & \multicolumn{3}{c}{Budget of 32,768 tokens} \\
\cmidrule(lr){2-5}\cmidrule(l){6-8}
 & GSM8K & MATH500 & AIME24 & AIME25 & AIME24 & AIME25 & MATH500 \\
\midrule
\multicolumn{8}{c}{\textit{\Tone, top-p 0.7}} \\
\midrule
LoopSpec / SGLang AR & 2.25 & 1.95 & 1.60 & 1.53 & 1.58 & 1.53 & 1.80 \\
\method / our AR & 3.04 & 3.20 & 2.99 & 2.96 & 3.05 & 3.15 & 3.46 \\
\methodH / our AR & \textbf{3.40} & \textbf{3.63} & \textbf{3.15} & \textbf{3.10} & \textbf{3.31} & \textbf{3.41} & \textbf{4.01} \\
\method / LoopSpec & 1.35 & 1.64 & 1.87 & 1.93 & 1.93 & 2.06 & 1.92 \\
\methodH / LoopSpec & \textbf{1.51} & \textbf{1.86} & \textbf{1.97} & \textbf{2.03} & \textbf{2.09} & \textbf{2.23} & \textbf{2.23} \\
LoopSpec in our engine / our AR & 2.13 & 2.02 & 1.76 & 1.73 & 1.70 & 1.68 & 1.93 \\
\method / LoopSpec in our engine & 1.43 & 1.58 & 1.70 & 1.71 & 1.79 & 1.88 & 1.79 \\
LoopSpec, $D$ & 0.3123 & 0.2964 & 0.3021 & 0.3041 & 0.2976 & 0.2999 & 0.2919 \\
\method, $D$ & 0.2467 & 0.2113 & 0.2043 & 0.2039 & 0.1939 & 0.1851 & 0.1950 \\
\midrule
\multicolumn{8}{c}{\textit{Greedy}} \\
\midrule
LoopSpec / SGLang AR & 2.64 & 2.16 & 1.82 & 1.79 & 1.67 & 1.67 & 1.78 \\
CopyLoop / SGLang AR & 2.74 & 2.20 & 1.97 & 1.83 & 1.78 & 1.77 & 1.80 \\
\method / our AR & 3.34 & 3.76 & 4.08 & 3.85 & 5.48 & 5.51 & 5.71 \\
\methodH / our AR & \textbf{3.71} & \textbf{4.22} & \textbf{4.63} & \textbf{4.61} & \textbf{5.88} & \textbf{5.77} & \textbf{5.92} \\
\method / LoopSpec & 1.27 & 1.74 & 2.24 & 2.15 & 3.28 & 3.30 & 3.21 \\
\methodH / LoopSpec & \textbf{1.41} & \textbf{1.95} & \textbf{2.54} & \textbf{2.58} & \textbf{3.52} & \textbf{3.46} & \textbf{3.33} \\
\method / CopyLoop & 1.22 & 1.71 & 2.07 & 2.10 & 3.08 & 3.11 & 3.17 \\
\bottomrule
\end{tabular}

%% file: tables/tab_huginn.tex
\begin{tabular}{@{}l cccccc@{}}
\toprule
 & GSM8K & \makecell{MATH-\\500} & BBH & \makecell{Human-\\Eval+} & MBPP+ & \makecell{Geo.\\mean} \\
\midrule
\multicolumn{7}{@{}l}{\textit{All prompts}} \\
\method & 4.40 & 10.55 & 6.16 & 9.98 & 7.90 & 7.42 \\
\methodH & \textbf{5.56} & \textbf{11.55} & \textbf{6.57} & \textbf{10.24} & \textbf{8.80} & \textbf{8.24} \\
CopyLoop & 4.14 & 6.46 & 4.83 & 7.61 & 6.34 & 5.74 \\
LoopSpec & 3.28 & 3.74 & 3.28 & 4.39 & 4.14 & 3.74 \\
\midrule
\multicolumn{7}{@{}l}{\textit{Without repetition loops}} \\
\method & 3.96 & 4.76 & 5.83 & 9.08 & 4.91 & 5.47 \\
\methodH & \textbf{5.06} & \textbf{5.98} & \textbf{6.28} & \textbf{9.44} & \textbf{6.00} & \textbf{6.40} \\
CopyLoop & 3.78 & 3.59 & 4.72 & 7.02 & 4.24 & 4.53 \\
LoopSpec & 3.14 & 3.06 & 3.23 & 4.33 & 3.60 & 3.45 \\
\bottomrule
\end{tabular}

%% file: tables/tab_context.tex
\begin{tabular}{@{}l cccccc c@{}}
\toprule
 & 1K & 2K & 4K & 8K & 16K & 32K & \makecell{Geo.\\mean} \\
\midrule
\multicolumn{8}{@{}l}{\textit{Over LoopSpec}} \\
Ouro-1.4B & 1.44 & 1.60 & 1.62 & 1.65 & 1.51 & 1.30 & 1.51 \\
Ouro-2.6B & 1.33 & 1.39 & 1.38 & 1.44 & 1.54 & 1.33 & 1.40 \\
Raven-Llama & 1.66 & 1.85 & 1.91 & 1.87 & 2.08 & 1.77 & 1.85 \\
\midrule
\multicolumn{8}{@{}l}{\textit{Over CopyLoop}} \\
Ouro-1.4B & 1.38 & 1.67 & 1.79 & 1.91 & 1.75 & 1.50 & 1.66 \\
Ouro-2.6B & 1.24 & 1.41 & 1.48 & 1.54 & 1.66 & 1.40 & 1.45 \\
Raven-Llama & 1.14 & 1.41 & 1.58 & 1.59 & 1.78 & 1.81 & 1.53 \\
\bottomrule
\end{tabular}